\pdfoutput=1

\documentclass[11pt]{article}

\usepackage[final]{EMNLP2023} 

\usepackage{times}
\usepackage{latexsym}

\usepackage[T1]{fontenc}

\usepackage[utf8]{inputenc}

\usepackage{microtype}

\usepackage{inconsolata}

\usepackage{graphicx}
\usepackage{multirow}
\usepackage[nolist,nohyperlinks]{acronym}
\usepackage{textcomp}
\usepackage{graphicx}

\newcommand{\CyrCapA}{%
  \begingroup\normalfont
  \includegraphics[height=\fontcharht\font`\B]{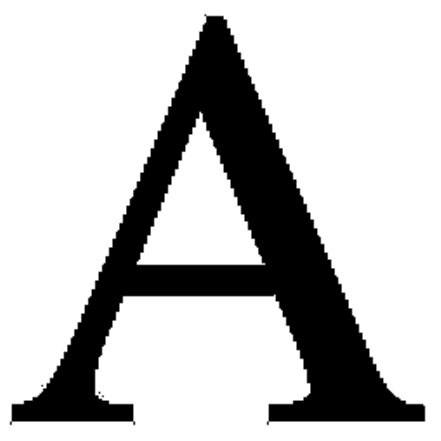}%
  \endgroup
}
\newcommand{\DevanagariDanda}{%
  \begingroup\normalfont
  \includegraphics[height=\fontcharht\font`\B]{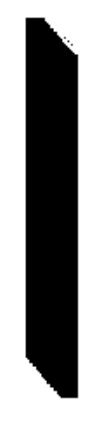}%
  \endgroup
}
\newcommand{\BackwardsC}{%
  \begingroup\normalfont
  \includegraphics[height=\fontcharht\font`\c]{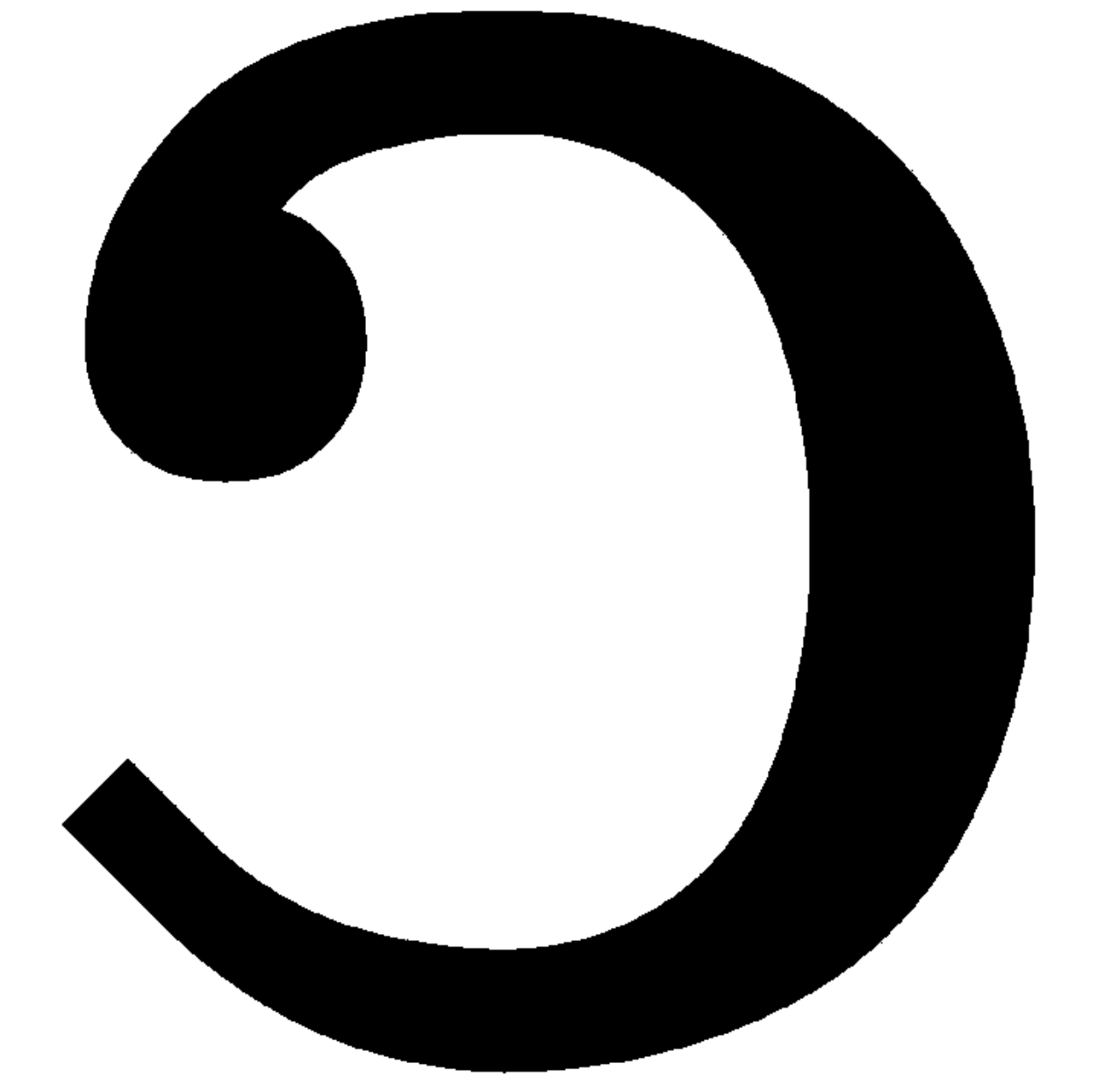}%
  \endgroup
}
\newcommand{\OpenO}{%
  \begingroup\normalfont
  \includegraphics[height=\fontcharht\font`\c]{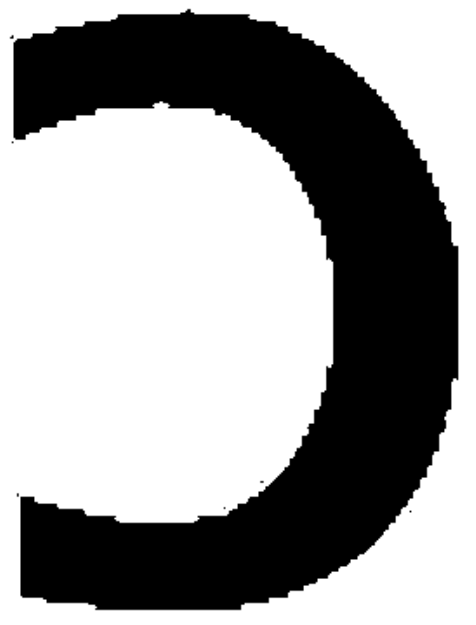}%
  \endgroup
}
\newcommand{\UnicodeFFFD}{%
  \begingroup\normalfont
  \includegraphics[height=\fontcharht\font`\B]{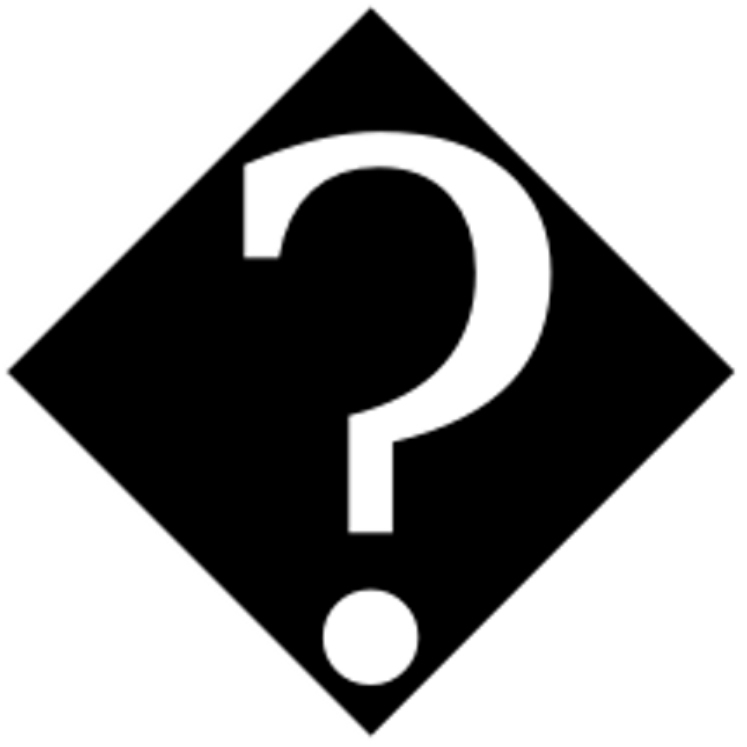}%
  \endgroup
}
\newcommand{\UnicodeFB}{%
  \begingroup\normalfont
  \includegraphics[height=\fontcharht\font`\B]{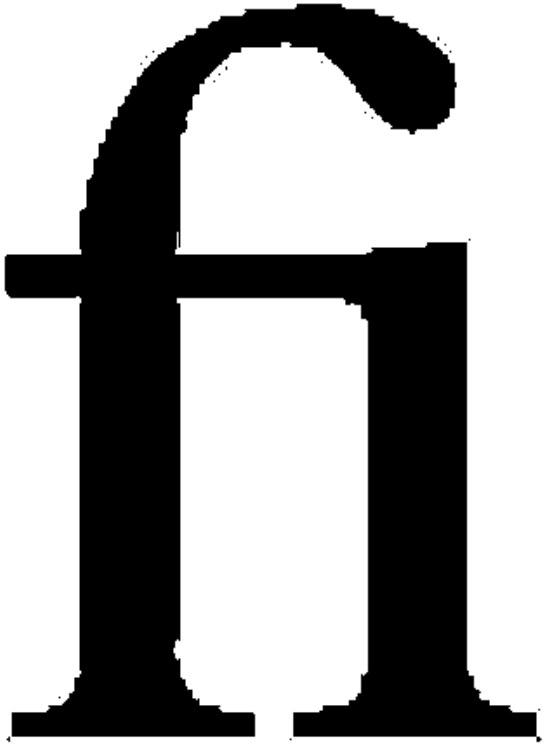}%
  \endgroup
}

\title{The eBible Corpus: Data and Model Benchmarks for Bible Translation for Low-Resource Languages}

\author{Vesa {\AA}kerman \\
  SIL International\\
  7500 W. Camp Wisdom Rd. \\
  Dallas, TX \\\And
  David Baines \\
  SIL International\\
  7500 W. Camp Wisdom Rd. \\
  Dallas, TX \\\And
  Damien Daspit \\
  SIL International\\
  7500 W. Camp Wisdom Rd. \\
  Dallas, TX \\\AND
  Ulf Hermjakob \\
  University of Southern California\\
  Information Sciences Institute\\
  4676 Admiralty Way \#1001 \\
  Marina del Rey, CA \\\And
  Taeho Jang \\
  Payap University Linguistics Department\\
  Super-highway Chiang Mai – Lumpang Road \\
  Amphur Muang, Chiang Mai, 50000 Thailand \\\AND
  Colin Leong \\
  University of Dayton\\
  300 College Park \\
  Dayton, OH \\\And
  Michael Martin \\
  SIL International\\
  7500 W. Camp Wisdom Rd. \\
  Dallas, TX \\
  \texttt{michael\_martin@sil.org} \\\And
  Joel Mathew \\
  University of Southern California\\
  Information Sciences Institute\\
  4676 Admiralty Way \#1001 \\
  Marina del Rey, CA \\\AND
  Jonathan Robie \\
  Clear Bible\\
  2990 Franklin Ave SW \#8\\
  Grandville, MI \\\And
  Marcus Schwarting \\
  University of Chicago\\
  Department of Computer Science\\
  5801 South Ellis Avenue\\
  Chicago, IL}

\begin{document}
\maketitle
\begin{abstract}
Efficiently and accurately translating a corpus into a low-resource language remains a challenge, regardless of the strategies employed, whether manual, automated, or a combination of the two.
Many Christian organizations are dedicated to the task of translating the Holy Bible into languages that lack a modern translation.
\Ac{BT} work is currently underway for over 3000 extremely low resource languages. We introduce the eBible corpus: a dataset containing 1009 translations of portions of the Bible with data in 833 different languages across 75 language families.
In addition to a \ac{BT} benchmarking dataset, we introduce model performance benchmarks built on the \ac{NLLB} \ac{NMT} models.
Finally, we describe several problems specific to the domain of \ac{BT} and consider how the established data and model benchmarks might be used for future translation efforts.
For a \ac{BT} task trained with \ac{NLLB}, Austronesian and Trans-New Guinea language families achieve 35.1 and 31.6 BLEU scores respectively, which spurs future innovations for \ac{NMT} for low-resource languages in Papua New Guinea.
\end{abstract}

\section{Introduction}
There has been significant progress recently towards solving multiple problems in the field of \ac{NLP}.
Most of these advances, however, are skewed towards languages of wider communication (LWCs).
Though there is ongoing work in low resource languages, the scarcity of training data and benchmarks for meaningful comparison of proposed techniques in such languages slow down the pace of research.

The Holy Bible has been translated to a very large number of languages of the world with continued work to modernize multiple translations.
Historically, \ac{BT}s have been foundational to the standardizing and revitalization of language for various communities.
Therefore, this data has the potential to be the starting point for \ac{NLP} research in many extremely low resource languages.
It would especially be useful for benchmarking model performance for \ac{NLP} researchers working in the Biblical domain against modern techniques.
Though not all translations are published under a permissive license for reuse, eBible.org has curated more than 1000 translations in various formats that are unencumbered.
Yet there are domain specific nuances and issues in the data format \cite{usfm}, structure (versification) and encoding that need careful handling and have been observed by the authors as an impedance to use the data efficiently for \ac{NLP}.

In this work we tackle the problem of scarcity of data and a model benchmark (for machine translation) in low resource languages by: 
Collecting, parsing and cleaning 1009 translations from eBible.org which have been automatically verified to be under a permissive license. These are made available as a verse(footnote on definition of verse)-wise parallel corpus across 833 languages.
Designing domain relevant benchmarking tasks that take into consideration the textual and stylistic variations in the content, having multiple related languages to a target language and the realities of the progress of a \ac{BT} project on the ground.

To our knowledge, this is the first time such a large unencumbered multilingual corpus and carefully designed benchmark has been released to the \ac{NLP} community.
This work draws heavily from our own experience and work with multiple recognized \ac{BT} teams, organizations and languages.

The following sections are organized so that we first briefly review existing relevant work on developing large Bible corpora and low resource machine translation.
We then detail the eBible corpus and its statistics along with the steps we took to parse and clean the data.
We share our experimental setup for the benchmark and the models we used.
We then share the experiment results and discuss findings to attract researchers to the issues faced in \ac{BT}.
Finally, we propose multiple research directions for future work using this dataset and provide concluding remarks.

\section{Background}
In this section we briefly review previous efforts to aggregate Biblical corpora.
We then consider previous \ac{NLP}-driven strategies for multilingual translation to such low-resource languages, including those specific to \ac{BT} tasks.

\subsection{Previously Aggregated Data}
The number of languages represented in Biblical corpora has been rising steadily over the last few decades.
Resnik et. al. produced a parallel corpus with 13 languages in 1999 \cite{resnik1999bible}, and by 2015 the corpus of Christodouloupoulos et. al. contained Bibles spanning 100 languages \cite{christodouloupoulos2015massively}.
In 2020, McCarthy et. al. reported on an effort to scrape and align Scriptures from various sources.
With over 1600 languages and 4000 unique translations represented it is most likely the largest Biblical corpus ever compiled, unfortunately this corpus is not publicly available \cite{mccarthy2020johns}.
Other online archives of open-license partial and full \ac{BT}s exist, but have not been made available in a format amenable to statistical or deep learning driven translation tasks.

Outside of \ac{BT}, the general problem of translating text into extremely low resource languages remains a challenge, primarily due to a lack of high-quality open-source data.
Datasets such as FLORES-101 \cite{goyal2022flores}, SALT \cite{akera2022machine} and AmericasNLI \cite{kann2022americasnli} previously provided a starting point for model training.
With the release of \ac{NLLB} also came the FLORES-200 dataset \cite{costa2022no}, which contains 3001 sentences sampled across 204 total languages.
FLORES-200 provides a many-to-many multilingual data benchmark which is the largest to date.

\subsection{Previous Translation Models}
We define a translation task as follows.
Suppose we are given a passage which is readily available in one or multiple source languages.
Suppose we also have a target language for which the passage has not been previously translated.
We define the translation task for this passage as the creation of a mapping between the source(s) and the target.
In the case of \ac{BT}, a mapping between passages is carried out by verse, but in the general case this can be performed by sentence.
The first non-classical machine translation models relied on \ac{SMT}, and include alignment-based strategies \cite{dyer2013simple}, Markovian methods \cite{deng2008hmm}, and many other frequentist and Bayesian approaches \cite{koehn2009statistical}.
Of particular interest is the work of \cite{wu2018creating}, which employs an \ac{SMT} approach on the Bible corpus compiled by \cite{mccarthy2020johns}.
\Ac{NMT} is a natural extension of \ac{SMT}, and utilizes a neural network architecture to train directly on source and target texts.
Basic \ac{NMT} implementations use an encoder and a decoder structure, and may forgo the recurrence and attention mechanism characteristic of transformers.
The OpenNMT package provides a turnkey implementation for fine-tuning \ac{NMT} models for specific translation tasks \cite{klein2017opennmt}.

Transformers designed for translation tasks can be considered an extension of earlier \ac{NMT} models through the inclusion of recurrent layers and an attention mechanism.
Many transformer architectures have been modified for translation, including fairseq \cite{ott2019fairseq} and BERT \cite{zhu2020incorporating}.
Of particular interest is the work of Leides, which uses a fairseq architecture trained on \ac{BT}s across 50 different languages and available for general-purpose \ac{BT} tasks \cite{liedes_fairseq}.
Finally, Meta’s \ac{NLLB} model represents the current state-of-the-art \ac{NMT} transformer, trained on the FLORES-200 dataset representing over 200 different languages \cite{costa2022no}.

\section{Methods}
In this section we describe the content of the eBible corpus and our pipeline for aggregating and preprocessing \ac{BT}s.
We also present summary statistics describing the contents of the eBible corpus.
We then detail several benchmark translation tasks we performed across eight language families within the corpus.
Finally, we describe the model architectures and scoring methods by which we will evaluate model translation performance.

\subsection{The eBible Corpus}
\begin{figure*}
    \centering
    \includegraphics[width=0.99\textwidth]{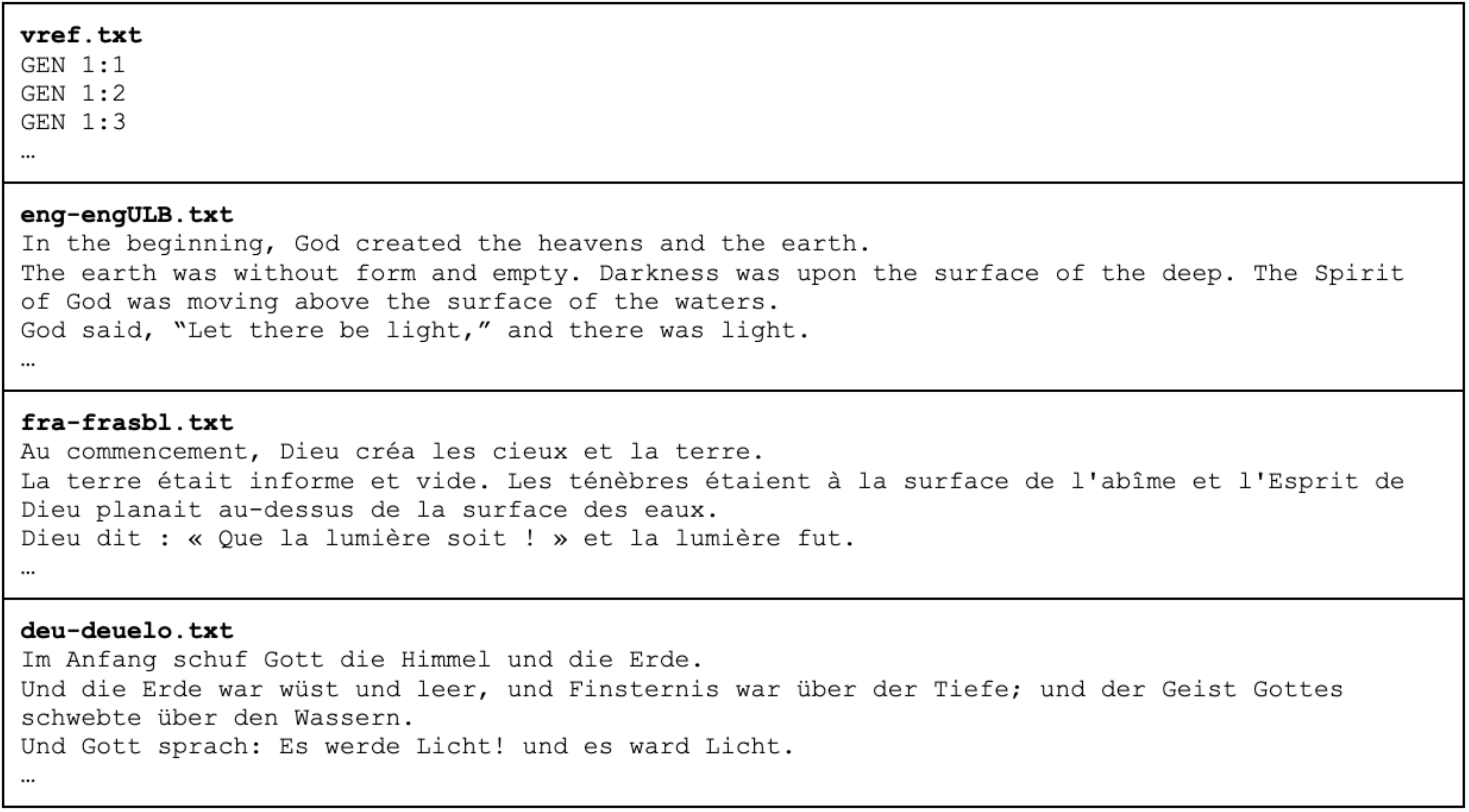}
    \caption{Sample verse extract files (vref, English, French, German).}
    \label{fig:text_extract}
\end{figure*}

We gathered and aligned 1,009 Scripture translations in 833 languages from eBible.org which are provided under a Creative Commons or similarly permissive licenses.
This includes 113 files under Attribution ShareAlike (CC BY-SA), two files under Attribution Non-Commercial (CC BY-NC), 106 files under Attribution No Derivs (CC BY-ND), 699 files under Attribution Non-Commercial No Derivs (CC BY-NC-ND), and 84 files under Public Domain.
After downloading these Bibles, the versification scheme (Original, English, Russian Orthodox, Russian Protestant, Septuagint, or Vulgate) for each Bible was inferred based on its content.
The text of each verse was extracted and all formatting, cross-references, footnotes, and other markup was removed.
The verse text was placed into an extract file with a verse-per-line format with 41,899 lines per file; the placement of each verse in the extract file was normalized to the Original versification scheme, allowing ready comparison of verses across translations.
A separate index file (\texttt{vref.txt}) records the verse reference for each line of all verse extract files.
Verse ranges were preserved by placing the verse range text on the first line of the range, and tagging subsequent lines from the verse range with the \texttt{<range>} indicator in the verse extract file. The corpus and code is openly available \cite{ebible_data}.
Additional tools used in this process include the SIL-NLP package \cite{silnlp} and the Wildebeest package \cite{wildebeest}.
Figure \ref{fig:text_extract} shows the general format of several extract files, designed to be easily ingested for \ac{NLP} analysis and machine translation tasks.

In order to reduce data fragmentation for \ac{NLP} tasks, we performed character-level cleaning on the eBible corpus using the Wildebeest tool, making 3.3 million changes to 220 out of the 1,009 \ac{BT}s, the vast majority of which are not or barely perceptible to the human reader.

Changes include bringing complex character sequences into conventional order (e.g. Devanagari primary character, nukta, vowel sign), preferring composed characters per Unicode standard; correcting some look-alike characters, e.g. mapping Latin A to Cyrillic \CyrCapA{} in Cyrillic-script text, or mapping Latin l to Devanagari danda \DevanagariDanda{} where appropriate; character normalization, e.g. reversed c to open o (\BackwardsC{} → \OpenO{}); for one translation, mapping the replacement character \UnicodeFFFD{} to open/close double/single quotes; correcting some comma errors (deleting spaces before a comma, adding a space after a comma, removing duplicate commas); decomposing some ligatures (e.g. \UnicodeFB{} → fi); and more\footnote{Wildebeest complex character order normalization generally matches the dominant forms of most original \ac{BT}s and other corpora; it is closest to Unicode’s NFC, but unlike NFC follows conventional order. E.g., NFC order of the above pattern is: Devanagari primary character, vowel sign, nukta.}.
Slightly over half of these changes were done fully automatically, using the Wildebeest Normalization script \texttt{wb\_normalize.py}; the remaining changes were made with an eBible specific script \texttt{wb\_bible\_plus.py} based on a manual review using the Wildebeest Analysis script \texttt{wb\_analysis.py}.
Some issues raised by the Wildebeest Analysis script have not been addressed, such as private-use characters in two \ac{BT}s (dwr-dwrENT, gof-gofENT), and some residual wrong-script characters that can’t readily be corrected automatically or semi-automatically.

\begin{figure}
    \centering
    \includegraphics[width=0.45\textwidth]{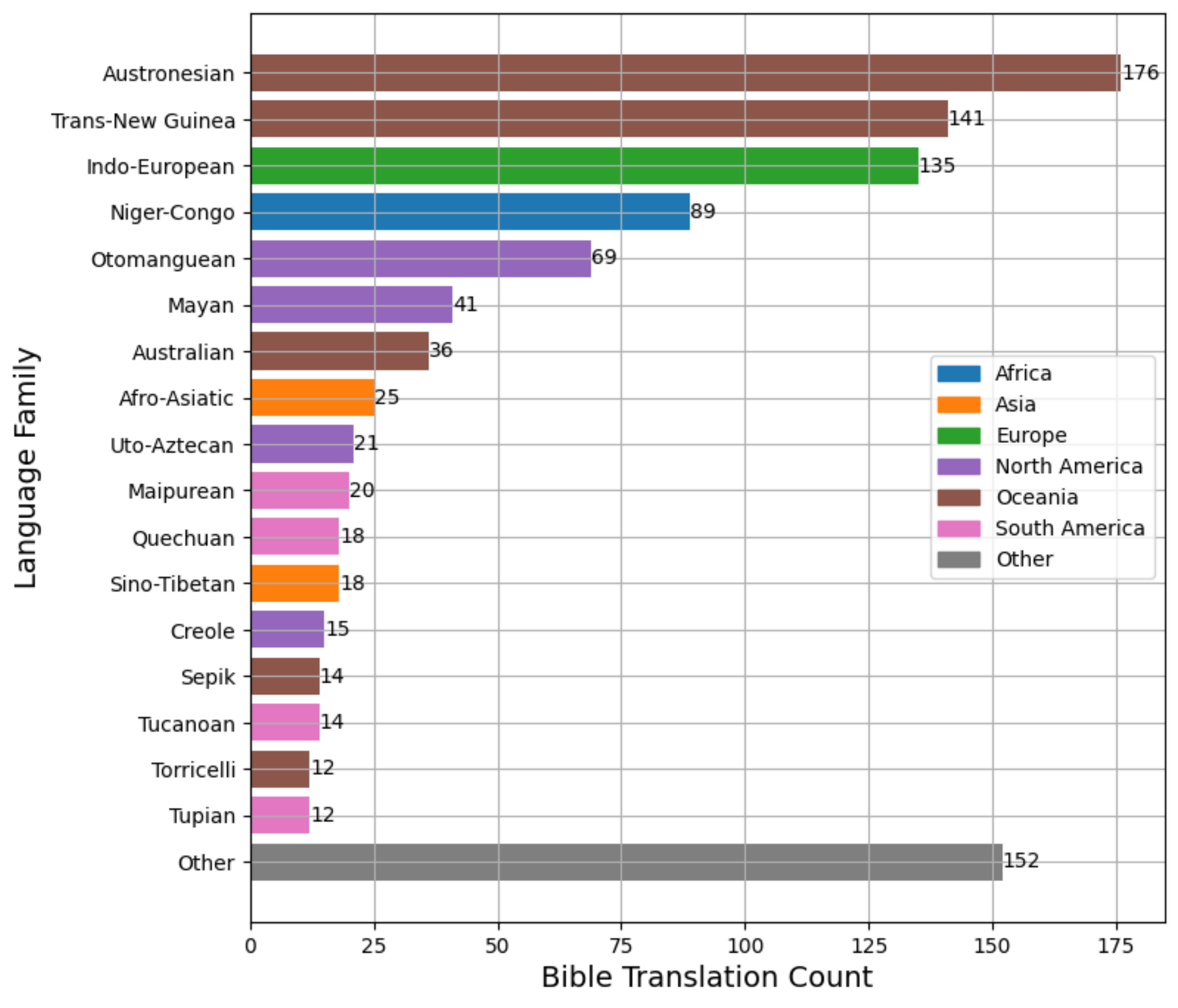}
    \caption{Count of \ac{BT}s by language family.}
    \label{fig:lang_family}
\end{figure}

\begin{figure}
    \centering
    \includegraphics[width=0.45\textwidth]{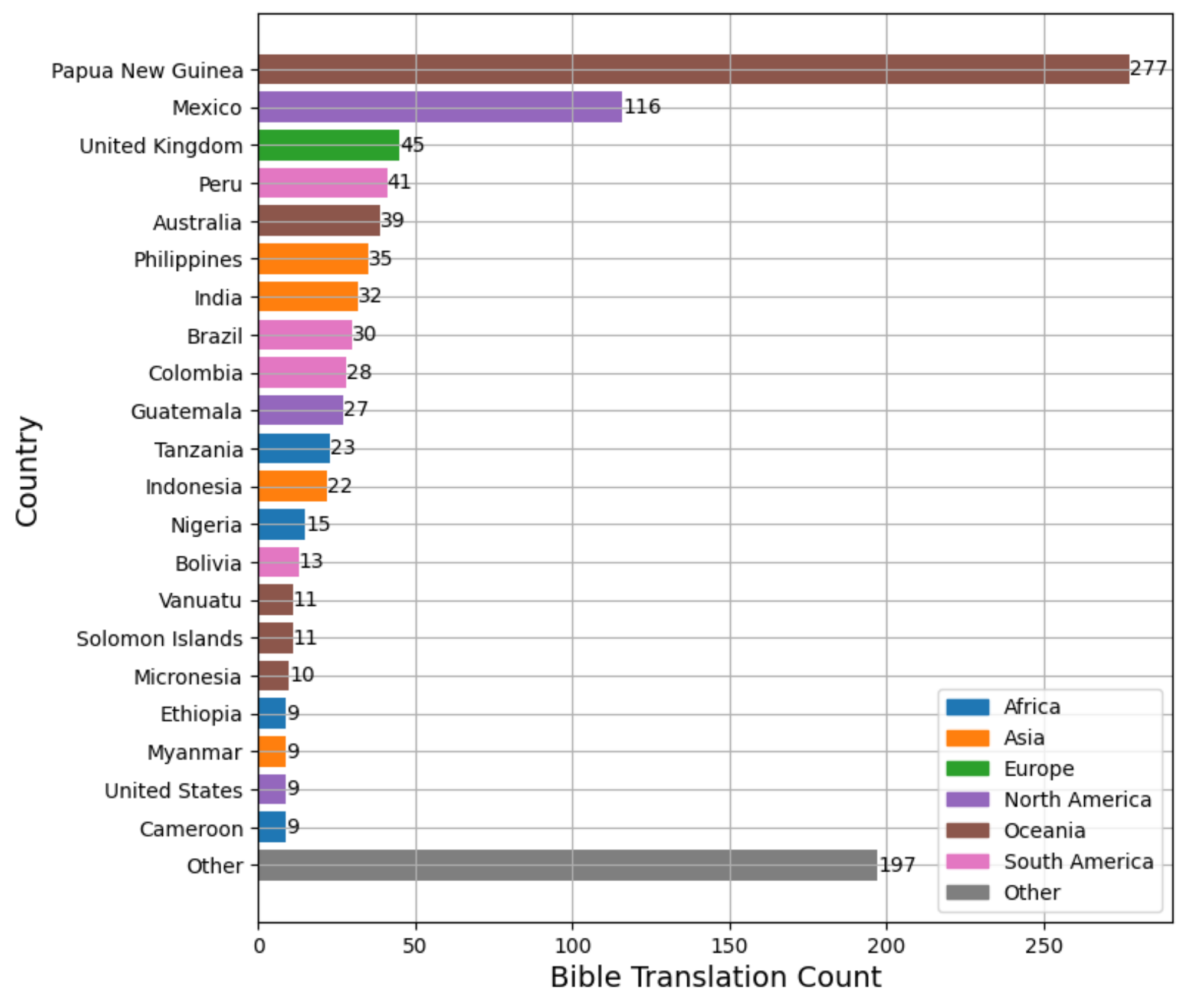}
    \caption{Count of \ac{BT}s by country.}
    \label{fig:country}
\end{figure}

\begin{figure}
    \centering
    \includegraphics[width=0.45\textwidth]{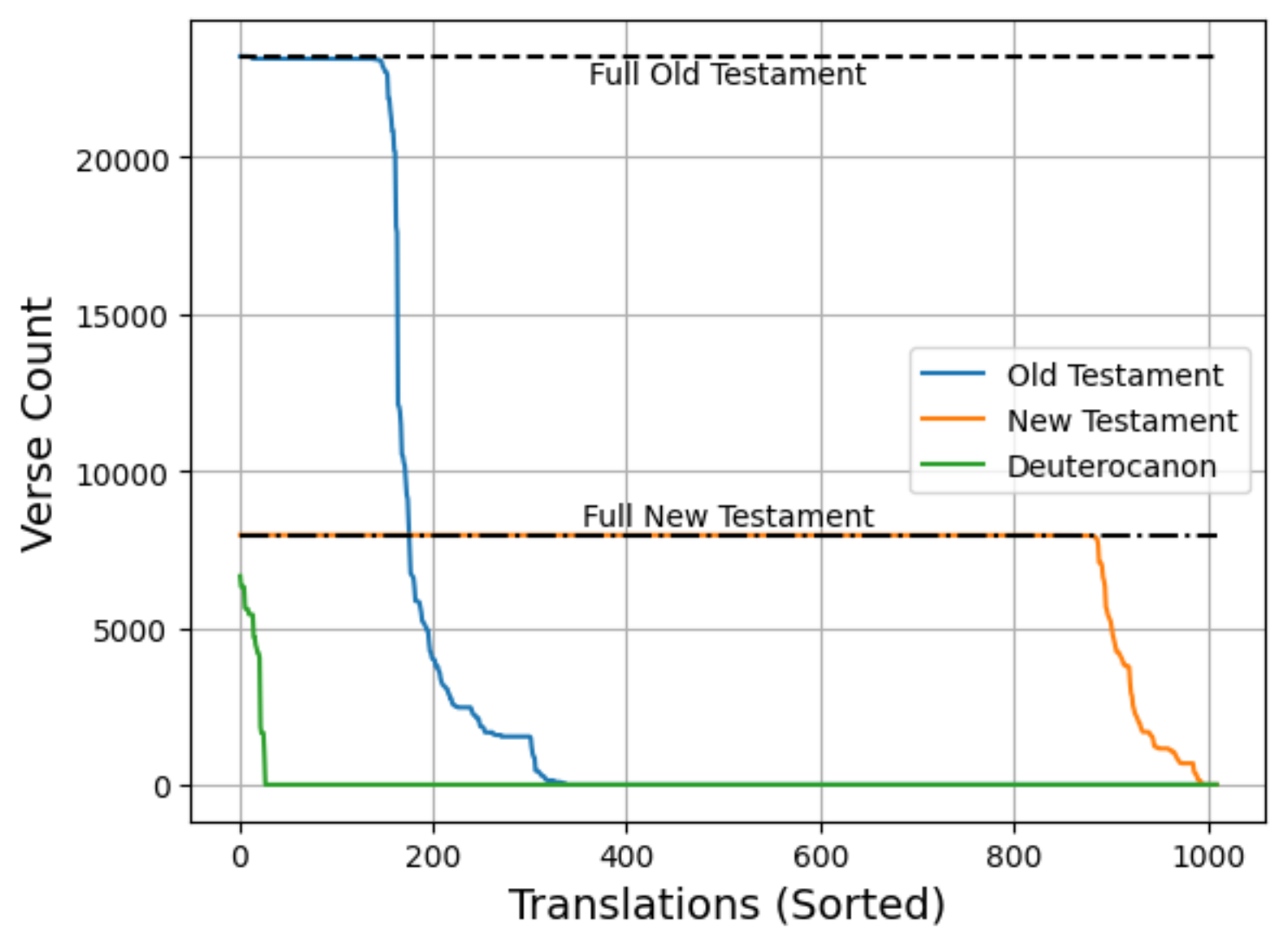}
    \caption{Sorted count of available verses per translation, separated by \ac{OT}, \ac{NT}, and  \ac{DT}.}
    \label{fig:verse_count}
\end{figure}

The eBible corpus exhibits a wide diversity of languages, as shown in Figure \ref{fig:lang_family} and Figure \ref{fig:country}.
Roughly a quarter of the translations are in languages spoken primarily in Papua New Guinea, which is widely known as the most linguistically diverse country in the world.
Many translations are in languages considered ultra-low-resource, and additional texts in some languages may not be readily available.
A large portion of the \ac{BT}s are partially complete, as shown in Figure \ref{fig:verse_count}.
Experts often start with translating the \ac{NT} before proceeding to the \ac{OT}, which is reflected in the availability of full \ac{NT} translations versus full \ac{OT} translations.
A small number of translations also include the Deuterocanon.
These are often translated last or not at all.
While they are included in the eBible corpus, we exclude them from further analysis based on their overall sparsity.

\subsection{Benchmark Translation Tasks}
In addition to a standard benchmark train/test/validation splitting of our paired-verse corpora, we also benchmark model performance based on more realistic translation approaches.
We chose several additional tasks motivated by plausible trajectories taken by those seeking to translate the Bible into a low-resource language.
For example, the progression of a translation may begin with the Gospels, then certain epistles, followed by portions of the Pentateuch, and so on.
In this fashion, we can train a model on content obtained earlier in the translation progression, then measure performance on content typically translated later.
Our translation tasks include:

\begin{itemize}
\item \emph{Randomized \ac{CV}} task.
Translation pairs are delimited by Bible verses available in both source and target corpora.
Due to the small size of the translation corpora (relative to those normally employed for \ac{NMT}), we do not use a standard 80\%/10\%/10\% split.
We instead set aside 250 randomly selected verses for testing and validation sets respectively, with the training set being the remaining verses, and running a five-fold \ac{CV} for scoring.
This task will represent an upper bound for possible model translation performance, with other tasks likely to perform worse overall.
\item \emph{Gospel Translation} task.
Train a model on the Gospel of Mark (MRK), test the model on the Gospel of Matthew (MAT).
We selected this task because Mark is often the first book to be translated, with Matthew to follow.
Some sections of Mark are also found in Matthew, however Mark is also a shorter book.
For three-letter Biblical book codes, readers are directed to Appendix C.
\item \emph{Epistles Translation} task.
Train a model on the Gospels (MAT, MRK, LUK, JHN) and Acts of the Apostles (ACT), test the model on the five epistles (1TH, 2TH, 1TI, 2TI, TIT, collectively abbreviated as 5T).
\item \emph{\Ac{NT} Completion} task.
Train a model on the entire \ac{NT} except Romans (ROM) and Revelation (REV), test the model on the books of Romans (ROM) and Revelation (REV).
Due to their translation difficulty, these books are often among the last books of the NT to be translated.
\item \emph{Early \Ac{OT} translation} task.
Train a model on the entire \ac{NT}, test the model on selected books of the \ac{OT} (GEN, EXO, LEV, NUM, DEU, RUT, PSA, JON, collectively abbreviated as Early \ac{OT}).
These books are often the first from the \ac{OT} to be translated.
\item \emph{Late \Ac{OT} translation task}.
Train a model on the entire Bible excluding minor prophets, test the model on \ac{OT} minor prophets (HOS, JOL, AMO, OBA, MIC, NAH, HAB, ZEP, HAG, ZEC, MAL).
These books were chosen as books that are typically among the last to be translated.
\item \emph{Related Language task}.
Train a model to translate from the source language into the target language and into a related language using the same train/test splits used for the \emph{Gospel Translation}, \emph{Epistles Translation}, and \emph{\Ac{NT} Completion tasks}.
This task is intended to explore the potential for improving translation accuracy through the use of other completed translations.
\end{itemize}

We focus on eight specific translation pairings with a translation source and target, spanning unique language families, as described in Table \ref{tab:lang_pairings}.
These translation pairings are selected to represent countries and language families with a significant number of active \ac{BT} projects and with reasonable representation in the eBible corpus.
Within each selected language family, source and target language translations were selected to simulate the work of a \ac{BT} team using a reference translation from a national or gateway language as a guide for translating into their target language, with access to a related language translation for further guidance.
In creating these source/target/related language translation pairings, the  target language translation was selected by identifying a language family branch with multiple translations in the corpus, preferably from the same country or in close geographic proximity \cite{collin2010ethnologue}.
Preference was given to branches containing more total languages, some with translations and some without.
Within this branch, preference was given to languages with a full \ac{BT}, or with a \Ac{NT} and partial \ac{OT} content.
The next step was identifying a translation from the corpus in a national or gateway language to act as the source language translation, with priority given to more recent translations from these languages.
Symmetric \ac{HMM} word alignment models were trained between each candidate national/gateway language translation and a small group of candidate target language translations; the source/target translation pairing with the highest overall word alignment score was then chosen.
Finally, symmetric \ac{HMM} word alignment models were trained between the target language translation and the available related language translations from the language family branch.
Generally, the translation with the highest overall alignment score was selected for the related language.
However, in some cases a related language translation with a particularly high alignment score was excluded, based on the assumption that it may have been translated as an adaptation or using some other less generalized translation methodology.
Detailed information on each translation pairing is available in Appendix C.

\begin{table*}[]
\centering
\resizebox{\textwidth}{!}{%
\begin{tabular}{|c|c|c|c|c|}
\hline
\textbf{Lang. Family(ISO-639-5)} &
  \textbf{Branch(es)} &
  \textbf{Purpose} &
  \textbf{Language(ISO-639-3)} &
  \textbf{Country} \\ \hline
\multirow{3}{*}{\begin{tabular}[c]{@{}c@{}}Afro-Asiatic \\ (afa)\end{tabular}} &
  \multirow{3}{*}{\begin{tabular}[c]{@{}c@{}}Chadic \\ (afa : cdc)\end{tabular}} &
  Source &
  Hausa (hau) &
  Nigeria \\ \cline{3-5} 
 &
   &
  Target &
  Dangaléat (daa) &
  Chad \\ \cline{3-5} 
 &
   &
  Related &
  Fulfulde, Western Niger (fuh) &
  Niger \\ \hline
\multirow{4}{*}{\begin{tabular}[c]{@{}c@{}}Austronesian \\ (map)\end{tabular}} &
  \multirow{4}{*}{\begin{tabular}[c]{@{}c@{}}Malayo-Polynesian, \\ Central Eastern Malayo-Poly., \\ Eastern Malayo-Polynesian\\ (map : poz : pqe)\end{tabular}} &
  Source &
  Kuanua (ksd) &
  Papua New Guinea \\ \cline{3-5} 
 &
   &
  Target &
  Kandas (kqw) &
  Papua New Guinea \\ \cline{3-5} 
 &
   &
  Related &
  Ramoaaina (rai) &
  Papua New Guinea \\
& 
   &
   &
   &
   \\ \hline
\multirow{3}{*}{\begin{tabular}[c]{@{}c@{}}Dravidian\\ (dra)\end{tabular}} &
  \multirow{3}{*}{N/A} &
  Source &
  Tamil (tam) &
  India \\ \cline{3-5} 
 &
   &
  Target &
  Malayalam (mal) &
  India \\ \cline{3-5} 
 &
   &
  Related &
  Kannada (kan) &
  India \\ \hline
\multirow{3}{*}{\begin{tabular}[c]{@{}c@{}}Indo-European\\ (ine)\end{tabular}} &
  \multirow{3}{*}{\begin{tabular}[c]{@{}c@{}}Indo-Iranian, \\ Indo-Aryan \\ (ine : iir : inc)\end{tabular}} &
  Source &
  Hindi (hin) &
  India \\ \cline{3-5} 
 &
   &
  Target &
  Eastern Panjabi (pan) &
  India \\ \cline{3-5} 
 &
   &
  Related &
  Gujarati (guj) &
  India \\ \hline
\multirow{3}{*}{\begin{tabular}[c]{@{}c@{}}Niger-Congo\\ (nic)\end{tabular}} &
  \multirow{3}{*}{\begin{tabular}[c]{@{}c@{}}Atlantic-Congo \\  (nic : alv)\end{tabular}} &
  Source &
  Swahili (swh) &
  Tanzania \\ \cline{3-5} 
 &
   &
  Target &
  Kwere (cwe) &
  Tanzania \\ \cline{3-5} 
 &
   &
  Related &
  Vidunda (vid) &
  Tanzania \\ \hline
\multirow{3}{*}{\begin{tabular}[c]{@{}c@{}}Otomanguean\\ (cai : omq)\end{tabular}} &
  \multirow{3}{*}{Eastern Otomanguean} &
  Source &
  Spanish (spa) &
  Spain \\ \cline{3-5} 
 &
   &
  Target &
  Zapotec, Tabaa (zat) &
  Mexico \\ \cline{3-5} 
 &
   &
  Related &
  Tapotec, Cajonos (zad) &
  Mexico \\ \hline
\multirow{3}{*}{\begin{tabular}[c]{@{}c@{}}Sino-Tibetan \\ (sit)\end{tabular}} &
  \multirow{3}{*}{N/A} &
  Source &
  Nepali (npi) &
  Nepal \\ \cline{3-5} 
 &
   &
  Target &
  Tamang, Eastern (taj) &
  Nepal \\ \cline{3-5} 
 &
   &
  Related &
  Limbu (lif) &
  Nepal \\ \hline
\multirow{3}{*}{\begin{tabular}[c]{@{}c@{}}Trans-New Guinea \\ (paa : ngf)\end{tabular}} &
  \multirow{3}{*}{N/A} &
  Source &
  Tok Pisin (tpi) &
  Papua New Guinea \\ \cline{3-5} 
 &
   &
  Target &
  Yopno (yut) &
  Papua New Guinea \\ \cline{3-5} 
 &
   &
  Related &
  Iyo (nca) &
  Papua New Guinea \\ \hline
\end{tabular}%
}
\caption{\label{tab:lang_pairings}
Language pairings for machine translation benchmarks.
}
\end{table*}

Our benchmarking tasks will be approached using four different models. First, we use a \ac{SMT} technique; namely, the “fast\_align” package from \cite{dyer2013simple} which uses a fast implementation IBM2 word alignment strategy.
Second, we perform training on the OpenNMT TransformerBase architecture from \cite{klein2017opennmt} with a SentencePiece unigram tokenization. Next, we perform fine-tuning on the “No Language Left Behind” (NLLB) model architecture from Meta \cite{costa2022no}, both the small version consisting of 600 million tunable parameters (\ac{NLLB}-600M) and the medium distilled version consisting of 1.3 billion tunable parameters (\ac{NLLB}-1.3B-distilled), which have pre-trained weights available from HuggingFace.

We scored models across translation tasks using three different metrics: BLEU \cite{papineni2002bleu}, Sentence Piece BLEU (spBLEU) \cite{goyal2022flores}, character 3-gram F-score (chrF3) \cite{popovic2015chrf}.
While BLEU and spBLEU scores are correlated to some extent, spBLEU more readily accounts for language variations in scripting and agglutination. Models were trained with an early stopping criteria of +0.1 BLEU over four checkpoints (1000 steps per checkpoint).
Models used a batch size of 16 with four gradient accumulation steps, 4000 warm-up steps, and label smoothing of 0.2.
For languages unknown to the NLLB tokenizer, we added a new language code as a special token to the tokenizer.
All models were trained and evaluated on an NVIDIA A-100 with 40 GB of available VRAM.
This hardware is sufficient for fine-tuning the small and medium size NLLB architectures.
For the random cross-validation task, we train five distinct models on different train/test/validation verse pair splits.
For all other tasks, only a single model of each type is trained with the identified splits.

\section{Results}
We divide our results into four sections according to tasks.
We first present results for a random cross-validation task, then results for tasks specific to translating \ac{NT} books, then results for tasks specific to translating OT books, and finally results for tasks specific to the use of a related language translation.
Further results are available for analysis in the ebible-experiments repository \cite{ebible_experiments}.

\subsection{Cross-Validation Translation Task}
\begin{figure*}
    \centering
    \includegraphics[width=0.99\textwidth]{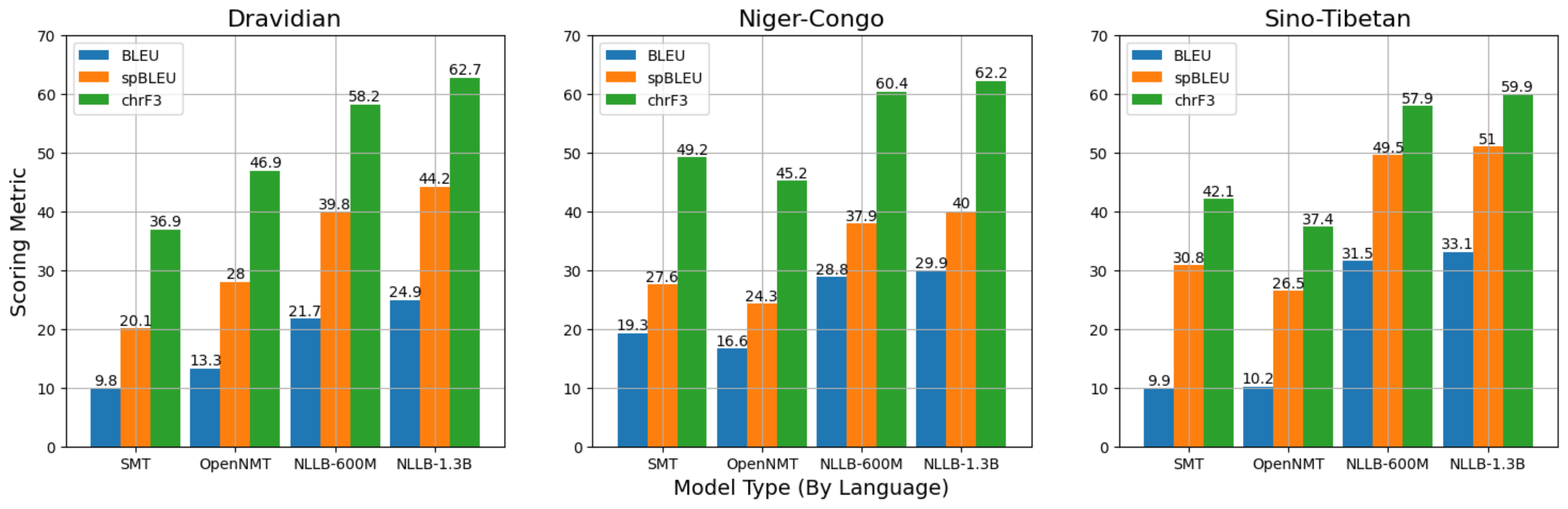}
    \caption{Bar chart of median BLEU, spBLEU, and chrF3 scores from \ac{SMT}, \ac{NMT}, and fine-tuned \ac{NLLB}-600M models for the Dravidian, Niger-Congo, and Sino-Tibetan translation pairings. Note that five-fold CV is not performed on the \ac{NLLB}-1.3B-distilled model due to training overheads.}
    \label{fig:model_comp}
\end{figure*}

We first consider the benchmark task for \emph{\ac{CV}} by verse.
This effectively provides an upper bound on possible model performance on later translation tasks.
We run our five-fold \ac{CV} across various models and present the results in Figure \ref{fig:model_comp}.
For the three language families shown in Table \ref{tab:NT_tasks}, we see a clear increase in performance from \ac{SMT} to OpenNMT in nearly all tests, with NLLB-600M out-performing both techniques across all metrics.
Likewise, the fine-tuned \ac{NLLB}-1.3B-distilled architecture out-performs its smaller counterpart in all instances.

\begin{figure}
    \centering
    \includegraphics[width=0.45\textwidth]{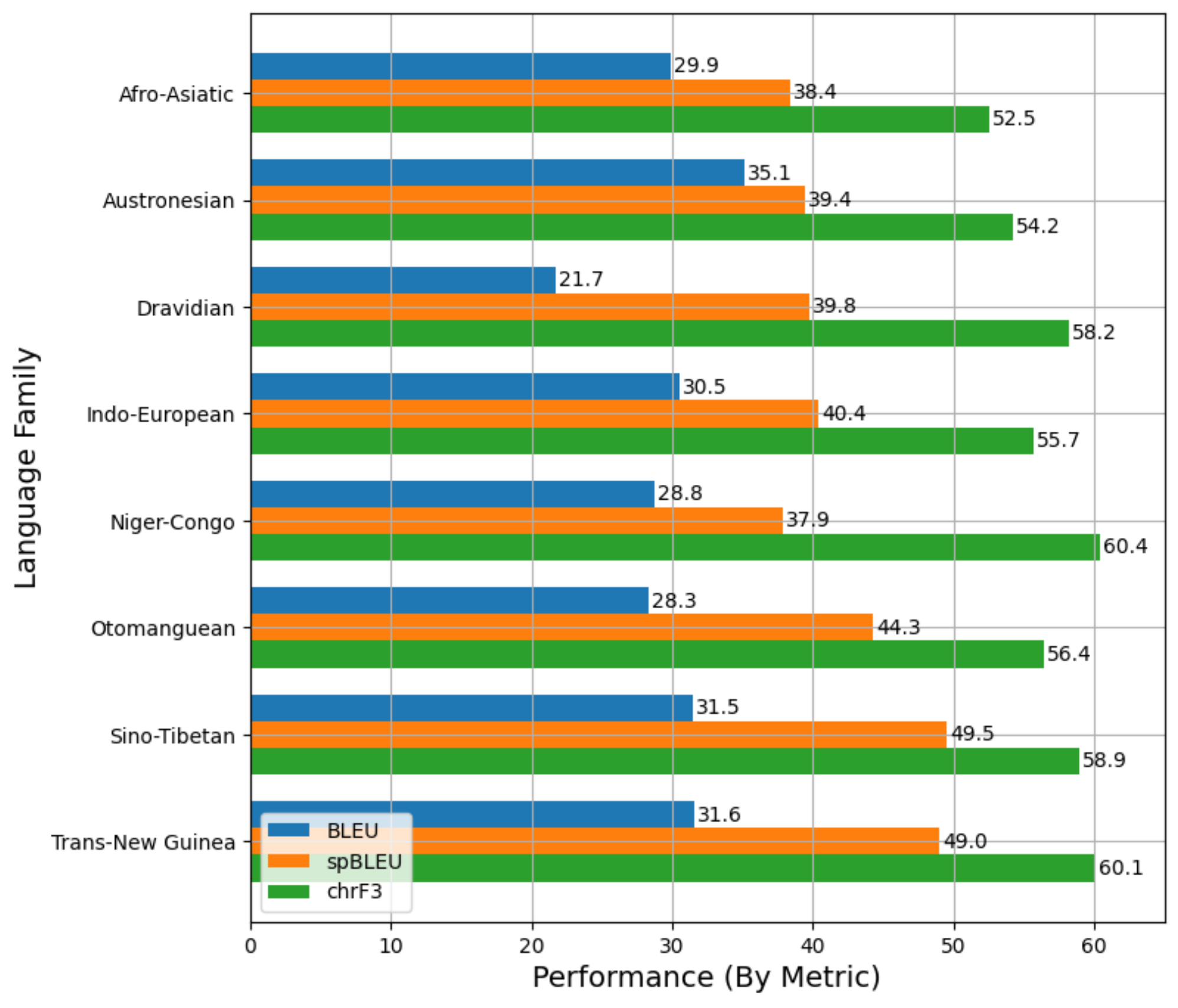}
    \caption{Bar chart of median BLEU, spBLEU, and chrF3 scores from a fine-tuned NLLB-600M model across eight language families for the CV task.}
    \label{fig:cv_task}
\end{figure}

We also present the test set results across all language families for the \emph{\ac{CV}} task using \ac{NLLB}-600M.
Figure \ref{fig:cv_task} shows a bar chart of the median BLEU, spBLEU, and chrF3 scores for all eight translation pairings.
Interestingly, we find no clear correlation between the scope (\ac{NT}-only, \ac{NT} with partial \ac{OT}, or full Bible) of the translation pairing and our selected scoring metrics.
The wide differences between word-level and subword-level metrics (BLEU and spBLEU) for some translation pairings such as Dravidian (+18.1), Sino-Tibetan (+18.0), and Trans-New Guinea (+17.4), compared to other translation pairings such as Austronesian (+4.3), highlight the value of examining a range of translation accuracy metrics when evaluating results on these benchmarks.

\subsection{New Testament Benchmark Tasks}
\begin{figure}
    \centering
    \includegraphics[width=0.45\textwidth]{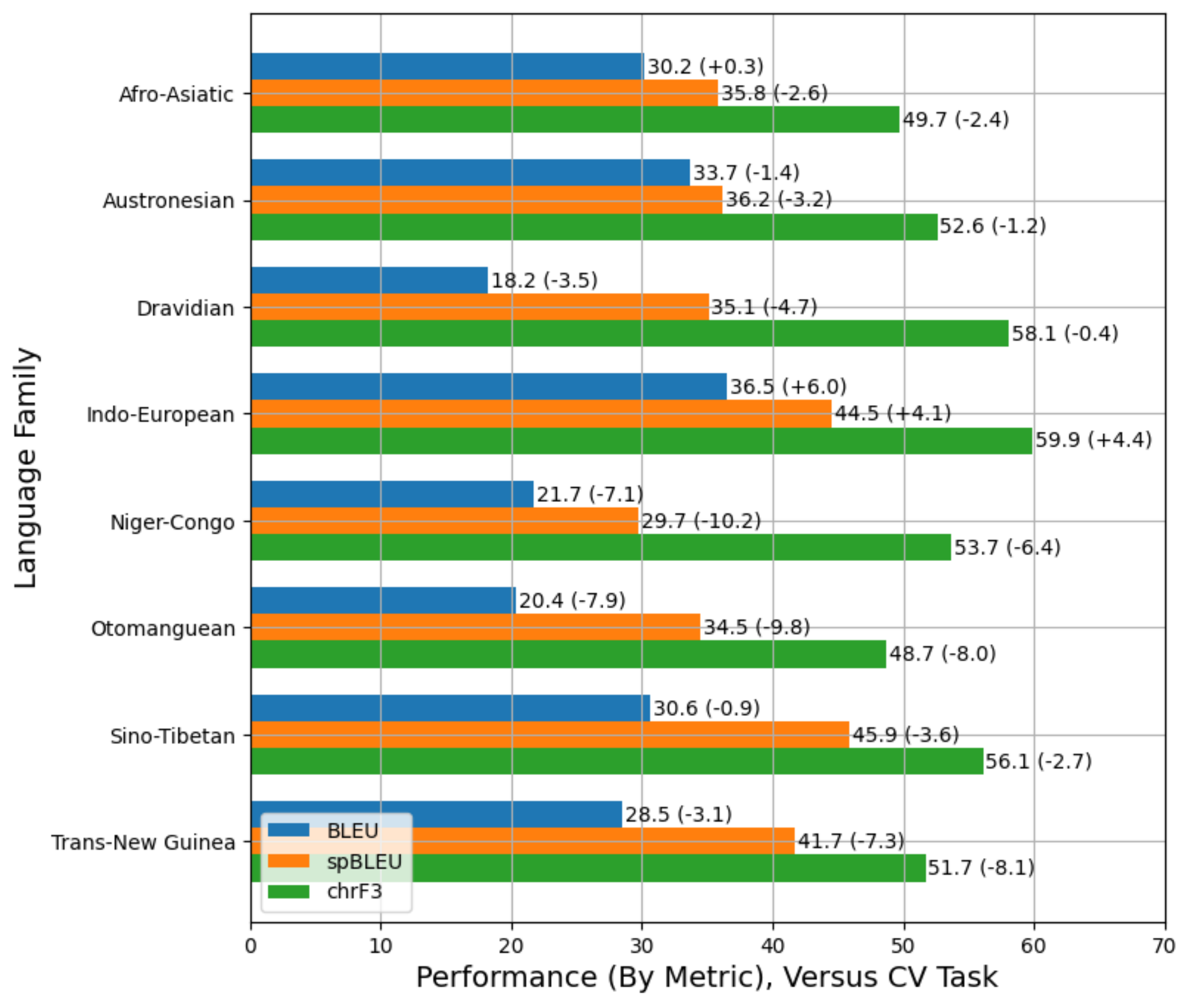}
    \caption{Bar chart of median BLEU, spBLEU, and chrF3 scores from a fine-tuned \ac{NLLB}-600M model across eight language families for the \emph{Gospel Translation task}, as compared to the \emph{\ac{CV}} task.}
    \label{fig:nt_benchmark}
\end{figure}

We first consider the \emph{Gospel Translation} task of fine-tuning an \ac{NLLB}-600M model using MRK as the training set and MAT as the test set.
Figure \ref{fig:nt_benchmark} shows the BLEU scores for this task across our eight language families, including a comparison to the \emph{\ac{CV}} task results.
We see that in general, there is a drop-off in performance compared to the \emph{\ac{CV}} task for six of the eight translation pairings.
This is attributed to the increased data heterogeneity and larger corpus used for training during the \emph{\ac{CV}} task.
These factors outweigh the benefits of the subject matter overlap between MRK and MAT.
However, results from these models for MAT are significantly better than their results for Epistles (see Table \ref{tab:NT_tasks}, Gospel Translation portion).

Next, we consider the \emph{Epistle Translation task} where \ac{NLLB}-600M models are either trained with MRK, or trained with the Gospels plus Acts of the Apostles, and then used to translate a selection of the Epistles (1TH, 2TH, 1TI, 2TI, and TIT).
These results are summarized in Table \ref{tab:NT_tasks}, which demonstrates that models perform significantly better when translating the selected Epistles when their training sets include these four additional books (MAT, LUK, JHN, ACT).

\begin{table*}[]
\centering
\resizebox{\textwidth}{!}{%
\begin{tabular}{|c|ccccc|ccccc|}
\hline
\multirow{2}{*}{\textbf{Translation Pairing}} &
  \multicolumn{5}{c|}{\textbf{Gospel Translation}} &
  \multicolumn{5}{c|}{\textbf{Epistle Translation}} \\ \cline{2-11} 
 &
  \multicolumn{1}{c|}{\textbf{1TH}} &
  \multicolumn{1}{c|}{\textbf{2TH}} &
  \multicolumn{1}{c|}{\textbf{1TI}} &
  \multicolumn{1}{c|}{\textbf{2TI}} &
  \textbf{TIT} &
  \multicolumn{1}{c|}{\textbf{1TH}} &
  \multicolumn{1}{c|}{\textbf{2TH}} &
  \multicolumn{1}{c|}{\textbf{1TI}} &
  \multicolumn{1}{c|}{\textbf{2TI}} &
  \textbf{TIT} \\ \hline
Afro-Asiatic &
  \multicolumn{1}{c|}{6.3} &
  \multicolumn{1}{c|}{5.5} &
  \multicolumn{1}{c|}{4.8} &
  \multicolumn{1}{c|}{6.9} &
  2.9 &
  \multicolumn{1}{c|}{14.8(+6.5)} &
  \multicolumn{1}{c|}{17.3(+11.8)} &
  \multicolumn{1}{c|}{11.3(+6.5)} &
  \multicolumn{1}{c|}{14.3(+7.4)} &
  9.3(+6.4) \\ \hline
Austronesian &
  \multicolumn{1}{c|}{12.7} &
  \multicolumn{1}{c|}{15.4} &
  \multicolumn{1}{c|}{13.3} &
  \multicolumn{1}{c|}{13.3} &
  13.9 &
  \multicolumn{1}{c|}{20.1(+7.4)} &
  \multicolumn{1}{c|}{21.5(+6.1)} &
  \multicolumn{1}{c|}{19.9(+6.6)} &
  \multicolumn{1}{c|}{23.5(+10.2)} &
  19.8(+5.9) \\ \hline
Dravidian &
  \multicolumn{1}{c|}{10.1} &
  \multicolumn{1}{c|}{12.0} &
  \multicolumn{1}{c|}{4.2} &
  \multicolumn{1}{c|}{7.4} &
  2.0 &
  \multicolumn{1}{c|}{11.4(+1.3)} &
  \multicolumn{1}{c|}{13.4(+1.4)} &
  \multicolumn{1}{c|}{7.0(+2.8)} &
  \multicolumn{1}{c|}{8.1(+0.7)} &
  5.2(+3.2) \\ \hline
Indo-European &
  \multicolumn{1}{c|}{24.1} &
  \multicolumn{1}{c|}{22.5} &
  \multicolumn{1}{c|}{17.1} &
  \multicolumn{1}{c|}{22.3} &
  15.5 &
  \multicolumn{1}{c|}{28.7(+4.6)} &
  \multicolumn{1}{c|}{26.2(+3.7)} &
  \multicolumn{1}{c|}{21.3(+4.2)} &
  \multicolumn{1}{c|}{26.7(+4.4)} &
  21.6(+6.1) \\ \hline
Niger-Congo &
  \multicolumn{1}{c|}{9.4} &
  \multicolumn{1}{c|}{13.4} &
  \multicolumn{1}{c|}{10.7} &
  \multicolumn{1}{c|}{10.8} &
  7.3 &
  \multicolumn{1}{c|}{14.9(+5.5)} &
  \multicolumn{1}{c|}{16.7(+3.3)} &
  \multicolumn{1}{c|}{15.1(+4.4)} &
  \multicolumn{1}{c|}{14.5(+3.7)} &
  13.6(+6.3) \\ \hline
Otomanguean &
  \multicolumn{1}{c|}{6.5} &
  \multicolumn{1}{c|}{7.4} &
  \multicolumn{1}{c|}{8.4} &
  \multicolumn{1}{c|}{8.4} &
  6.9 &
  \multicolumn{1}{c|}{16.2(+9.7)} &
  \multicolumn{1}{c|}{14.0(+6.6)} &
  \multicolumn{1}{c|}{15.5(+7.1)} &
  \multicolumn{1}{c|}{13.1(+4.7)} &
  15.6(+8.7) \\ \hline
Sino-Tibetan &
  \multicolumn{1}{c|}{13.2} &
  \multicolumn{1}{c|}{10.6} &
  \multicolumn{1}{c|}{9.4} &
  \multicolumn{1}{c|}{10.3} &
  8.7 &
  \multicolumn{1}{c|}{18.8(+5.6)} &
  \multicolumn{1}{c|}{17.2(+6.6)} &
  \multicolumn{1}{c|}{16.2(+6.8)} &
  \multicolumn{1}{c|}{18.2(+7.9)} &
  18.3(+9.6) \\ \hline
Trans-New Guinea &
  \multicolumn{1}{c|}{6.9} &
  \multicolumn{1}{c|}{8.4} &
  \multicolumn{1}{c|}{7.2} &
  \multicolumn{1}{c|}{7.6} &
  6.5 &
  \multicolumn{1}{c|}{18.7(+11.8)} &
  \multicolumn{1}{c|}{20.5(+12.1)} &
  \multicolumn{1}{c|}{16.9(+9.7)} &
  \multicolumn{1}{c|}{17.6(+10.0)} &
  15.0(+8.5) \\ \hline
\end{tabular}%
}
\caption{\label{tab:NT_tasks}
BLEU scores for \ac{NLLB}-600M fine-tuned models: \emph{Gospel Translation} and \emph{Epistle Translation} tasks.
}
\end{table*}

Finally, we consider the \emph{\ac{NT} Completion} task using a fine-tuned \ac{NLLB}-600M model.
We fine-tune models first on MRK only, then on the Gospels plus ACT, then on the entire \ac{NT} except Romans (ROM) and Revelation (REV).
In this fashion we form training sets with corpora of increasing size and literary breadth.
For each translation pairing, BLEU scores for both ROM and REV increase across these three tasks, indicating that the increased size and literary breadth of the training set is beneficial for the translation of these challenging \ac{NT} books.

\begin{table}[]
\centering
\resizebox{0.45\textwidth}{!}{%
\begin{tabular}{|c|cc|cc|cc|}
\hline
\multirow{2}{*}{\begin{tabular}[c]{@{}c@{}}Translation\\ Pairing\end{tabular}} &
  \multicolumn{2}{c|}{\textbf{\begin{tabular}[c]{@{}c@{}}Gospel\\ Translation\end{tabular}}} &
  \multicolumn{2}{c|}{\textbf{\begin{tabular}[c]{@{}c@{}}Epistle Translation\\ (vs Gospel Translation)\end{tabular}}} &
  \multicolumn{2}{c|}{\textbf{\begin{tabular}[c]{@{}c@{}}NT Completion\\ (vs Epistle Translation)\end{tabular}}} \\ \cline{2-7} 
                 & \multicolumn{1}{c|}{\textbf{ROM}} & \textbf{REV} & \multicolumn{1}{c|}{\textbf{ROM}} & \textbf{REV} & \multicolumn{1}{c|}{\textbf{ROM}} & \textbf{REV} \\ \hline
Afro-Asiatic     & \multicolumn{1}{c|}{7.0}          & 7.6          & \multicolumn{1}{c|}{14.1 (+7.1)}  & 15.1 (+7.5)  & \multicolumn{1}{c|}{18.9 (+4.8)}  & 16.0 (+0.9)  \\ \hline
Austronesian     & \multicolumn{1}{c|}{11.5}         & 15.3         & \multicolumn{1}{c|}{18.1 (+6.6)}  & 26.2 (+10.9) & \multicolumn{1}{c|}{23.2 (+5.1)}  & 28.6 (+2.4)  \\ \hline
Dravidian        & \multicolumn{1}{c|}{8.2}          & 9.7          & \multicolumn{1}{c|}{10.9 (+2.7)}  & 13.0 (+3.3)  & \multicolumn{1}{c|}{12.8 (+1.9)}  & 13.7 (+0.7)  \\ \hline
Indo-European    & \multicolumn{1}{c|}{23.2}         & 21.7         & \multicolumn{1}{c|}{26.5 (+3.3)}  & 26.5 (+4.8)  & \multicolumn{1}{c|}{30.0 (+3.5)}  & 29.2 (2.7)   \\ \hline
Niger-Congo      & \multicolumn{1}{c|}{11.4}         & 11.4         & \multicolumn{1}{c|}{16.9 (+5.5)}  & 20.3 (+8.9)  & \multicolumn{1}{c|}{20.1 (+3.2)}  & 23.5 (+3.2)  \\ \hline
Otomanguean      & \multicolumn{1}{c|}{7.6}          & 9.8          & \multicolumn{1}{c|}{15.2 (+7.6)}  & 19.0 (+10.2) & \multicolumn{1}{c|}{22.1 (+6.9)}  & 20.6 (+1.6)  \\ \hline
Sino-Tibetan     & \multicolumn{1}{c|}{12.0}         & 13.5         & \multicolumn{1}{c|}{19.5 (+7.5)}  & 21.2 (+7.7)  & \multicolumn{1}{c|}{23.3 (+3.8)}  & 23.3 (+2.1)  \\ \hline
Trans-New Guinea & \multicolumn{1}{c|}{8.8}          & 7.4          & \multicolumn{1}{c|}{18.5 (+9.7)}  & 18.3 (+10.9) & \multicolumn{1}{c|}{24.9 (+6.4)}  & 22.1 (+3.8)  \\ \hline
\end{tabular}%
}
\caption{\label{tab:NT_across_tasks}
BLEU scores for NLLB-600M fine-tuned models: \emph{Gospel Translation}, \emph{Epistle Translation}, \emph{\ac{NT} Completion} tasks.
}
\end{table}

\subsection{Old Testament Benchmark Tasks}
\begin{figure}
    \centering
    \includegraphics[width=0.45\textwidth]{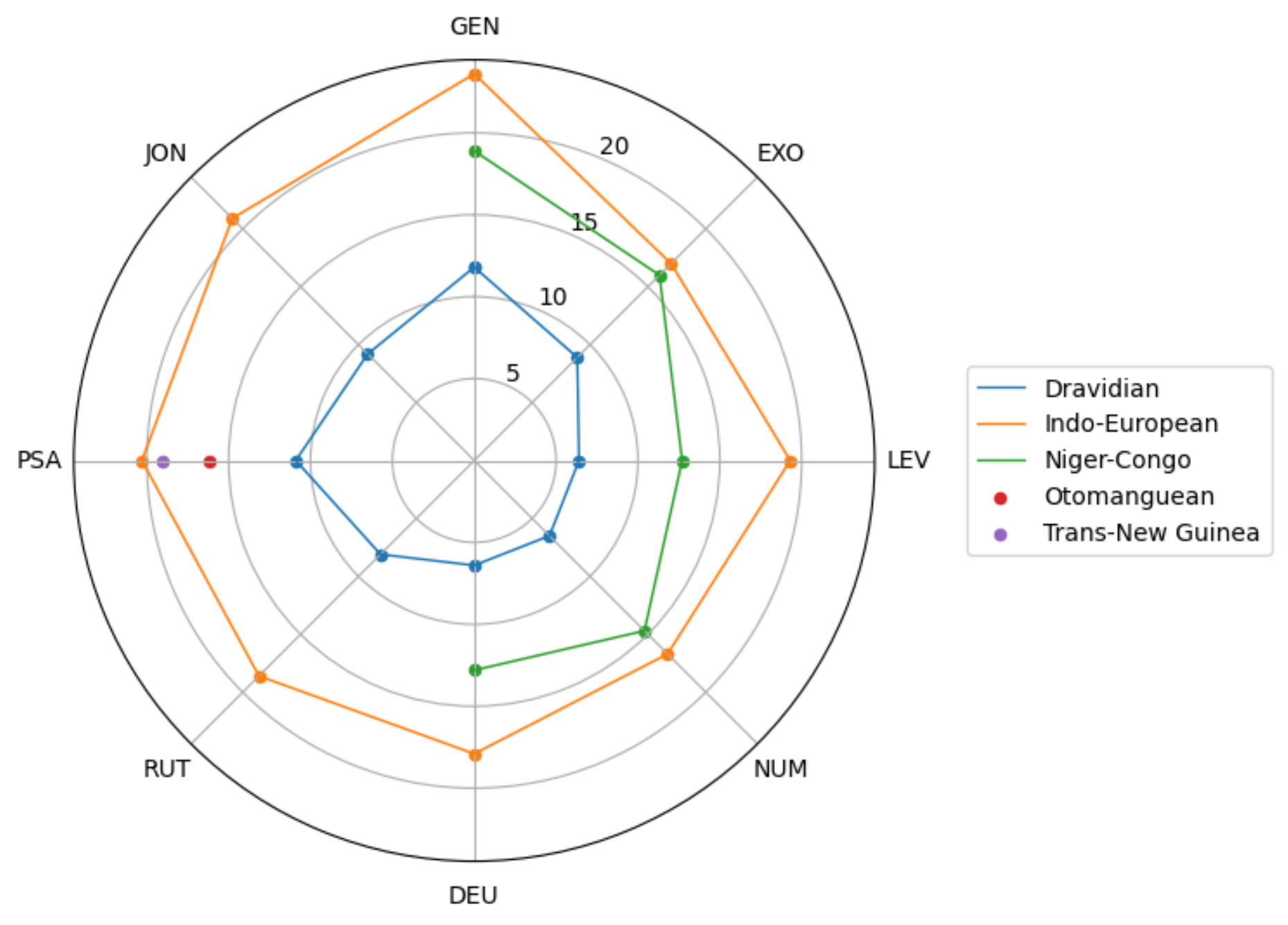}
    \caption{Radar plot of \ac{NLLB}-600M BLEU scores for four language families for the \emph{Early \ac{OT}} task. Missing points for Niger-Congo and Otomanguean indicate that these books are not included in the target language translation.}
    \label{fig:early_ot}
\end{figure}

For the \emph{Early \ac{OT}} translation task, the entire \ac{NT} is used as the training set; the fine-tuned model is then evaluated on various books that are typically translated early in an \ac{OT} translation project (GEN-DEU, RUT, PSA, JON).
Figure \ref{fig:early_ot} shows the BLEU scores of \ac{NLLB}-600M models trained across five translation pairs on the \ac{NT}, with translations performed across books in the Early \ac{OT} as the test set.
Empty cells indicate a lack of an available translation for a particular book in the target language.
For the tested Early \ac{OT} books, the highest BLEU scores within a translation pair are observed for GEN, while the lowest scores are observed for LEV, NUM, or DEU.

\begin{figure}
    \centering
    \includegraphics[width=0.45\textwidth]{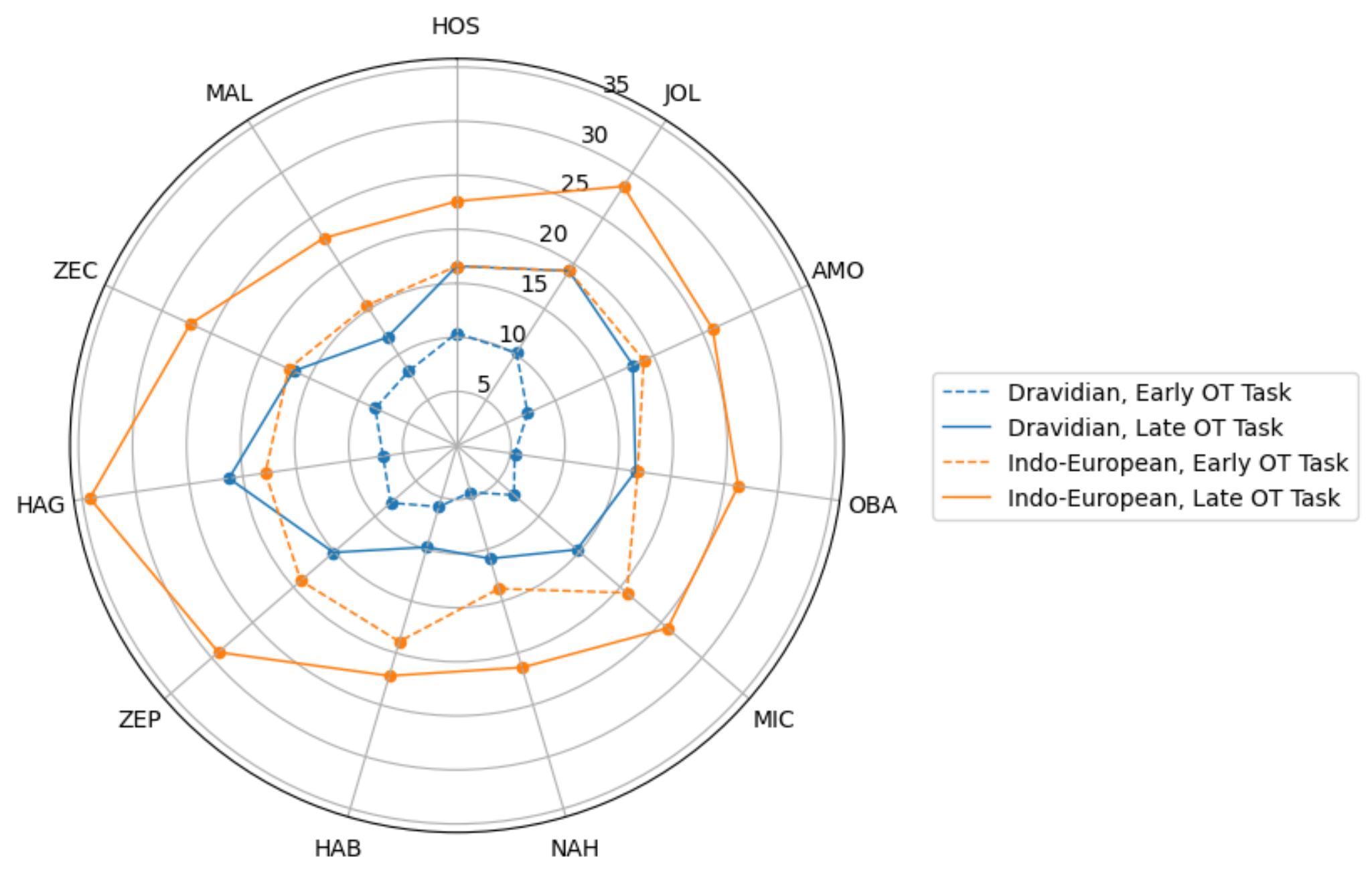}
    \caption{Radar plot of \ac{NLLB}-600M BLEU scores for Dravidian and Indo-European for \emph{Early} and \emph{Late \ac{OT}} tasks.}
    \label{fig:late_ot}
\end{figure}

Next, we assess the performance of \ac{NLLB}-600M on the \emph{Late \ac{OT}} translation task.
The books in this test set (the minor prophets) tend to be among the last books of the Bible to be translated.
Figure \ref{fig:late_ot} shows the BLEU scores for \ac{OT} minor prophet books across Dravidian and Indo-European language families, both in the case of models trained on the entire \ac{NT} and models trained on the entire Bible (excluding the minor prophets).
As before, a larger corpus of training data led to a large improvement in BLEU scores.

\subsection{Related Language Benchmark Tasks}
For the \emph{Related Language} task, an additional translation is selected from the eBible corpus for each of the eight language families.
\Ac{HMM} word alignment models are trained between the target language and each related language translation in the eBible corpus from the same branch of the language family; the translation with the best alignment to the target language is selected as the related language translation.
Then, the \emph{Gospel Translation}, \emph{Epistle Translation}, and \emph{\ac{NT} Completion} tasks are repeated using both the target language and the related language on the target side of the model.
For each task, the related language training data included the same verse pairs used for the target language training data; additionally, the related language training data included the verse pairs from the target language test set.

\begin{figure*}
    \centering
    \includegraphics[width=0.99\textwidth]{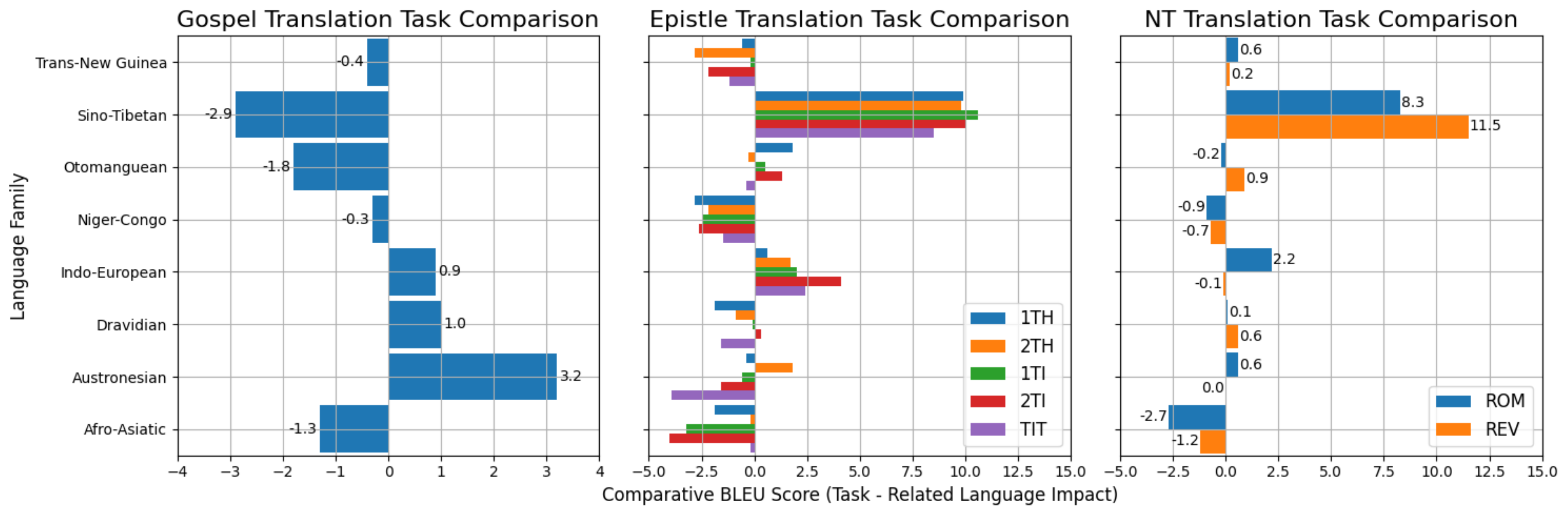}
    \caption{Impact of a Related Language Translation on the \emph{Gospel Translation}, \emph{Epistle Translation}, and \emph{NT Completion} tasks.}
    \label{fig:task_comparison_NT}
\end{figure*}

Figure \ref{fig:task_comparison_NT} compares the BLEU score deltas for the Related Language version of each task compared to the original version of the task.
BLEU score deltas across the eight language families are widely divergent, ranging from -2.9 to +3.2 BLEU (\emph{Gospel Translation} task), -4.0 to +10.6 BLEU (\emph{Epistle Translation} task), and -2.7 to +11.5 BLEU (\emph{\ac{NT} Completion} task).
Results for the Austronesian and Niger-Congo language families are strongest, while results for the Afro-Asiatic, Indo-European, Sino-Tibetan and Trans-New Guinea language families are the weakest.
These BLEU score deltas correlate well with the \ac{HMM} alignment scores between the target language and related language, with the Austronesian (0.68) and Niger-Congo (0.49) translation pairings exhibiting the highest alignment, and the Afro-Asiatic (0.21), Indo-European (0.30), Sino-Tibetan (0.27) and Trans-New Guinea (0.27) exhibiting the lowest alignment among the selected translation pairings.

\section{Discussion}
On the \emph{\ac{CV}} benchmark task, median accuracy metrics across the eight selected language families varied widely, ranging from 21.7 - 35.1 (BLEU), 37.9 - 49.5 (spBLEU), and 52.5 - 60.4 (chrF3).
No individual characteristic of the selected languages and translations correlated closely with the distribution of these results for any of these three accuracy metrics.
For instance, the scope of the source / target translation pairings varied from NT-only, to NT with OT portions, to full \ac{BT}s; while the highest BLEU score was seen for an \ac{NT}-only pairing (Austronesian (35.1)) and the lowest BLEU score was seen for a full Bible pairing (Dravidian (21.7)), other \ac{NT}-only pairings (Afro-Asiatic (29.9); Sino-Tibetan (31.5)) and full Bible pairings (Indo-European (30.5)) scored comparably.
Similar diversity was seen when comparing spBLEU and chrF3 scores to the scope of the translation pairings.
\ac{HMM} word alignment scores for the source/target translation pairings also do not correlate closely with the accuracy metrics on this benchmark; translation pairings with the highest (Niger-Congo (0.44)) and lowest (Otomanguean (0.19)) word alignment scores resulted in comparable BLEU scores (28.8 and 28.3, respectively).

\Ac{NLLB} characterized the resource level of each supported language as either high or low, with low resource languages being trained on less than 1M bitexts.
In our CV benchmark, the source languages were a mix of high resource (Hindi, Swahili, Spanish), low resource (Hausa, Tamil, Nepali, Tok Pisin), and unsupported (Kuanua).
However, there was no clear correlation between the NLLB resource level of the source language and the resulting metrics for the translation pairing.
Similarly, while the best chrF3 results were seen for translation pairings with languages using the Latin script (Niger-Congo (60.4); Trans-New Guinea (60.1)), which is well-represented in the NLLB vocabulary, other Latin script translation pairings performed relatively poorly (Afro-Asiatic (52.5); Austronesian (54.2)).  

While BLEU is a widely used, language agnostic metric for assessing machine translation accuracy, the fact that it is a word-level metric means that it can be difficult to interpret the metric across languages, particularly when attempting to judge the usefulness of a translation model for a less well known language.  Combining a word-level metric (e.g., BLEU) with a subword-level  metric (e.g., spBLEU) and a character-level metric (e.g., chrF3) provides a more nuanced view.

Table \ref{tab:NT_across_tasks} presents several sample predictions from the \emph{\ac{CV}} task with median BLEU, spBLEU, and chrF3 scores for their respective model\footnote{For the sake of simplicity, we are comparing predictions to existing translations.
This has obvious limitations.
For instance, translations may use different translation strategies, or there may even be mistakes in a translation.
Reading each of these texts in the target languages and creating a corrected text in order to compute a BLEU score is, however, beyond the scope of this paper.}.
The predictions are color-coded at the word level to give a general sense of the accuracy of each prediction, and suggest that, although the \ac{NLLB} models represent a strong improvement over earlier \ac{NMT} approaches, further improvement is needed.

\begin{figure}
    \centering
    \includegraphics[width=0.45\textwidth]{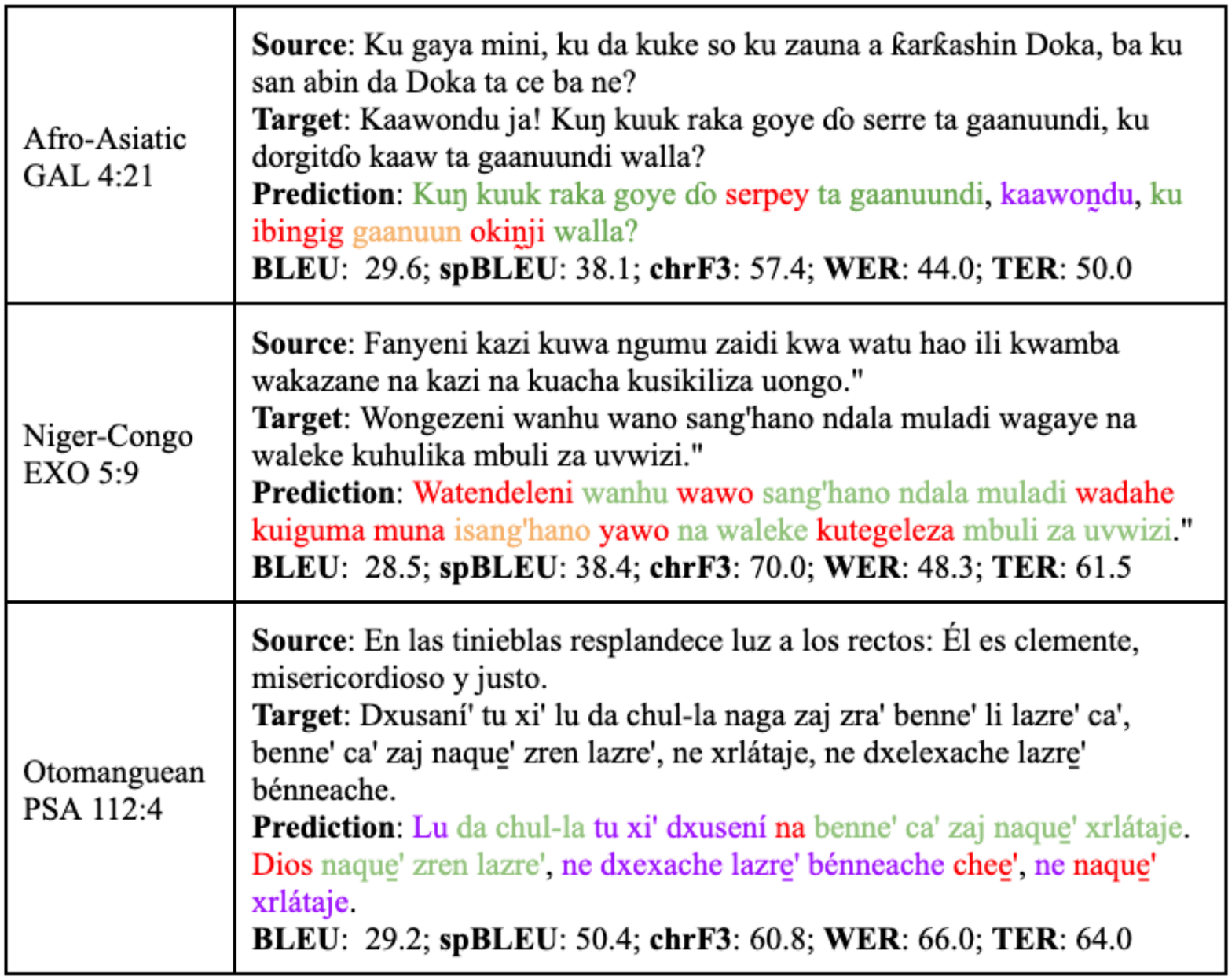}
    \caption{Sample predictions with median verse accuracy, \emph{\ac{CV}} task models.
    Green indicates correct words, yellow indicates partially correct words, purple indicates moved words, and red indicates incorrect words.}
    \label{fig:task_comparison}
\end{figure}

\begin{figure}
    \centering
    \includegraphics[width=0.45\textwidth]{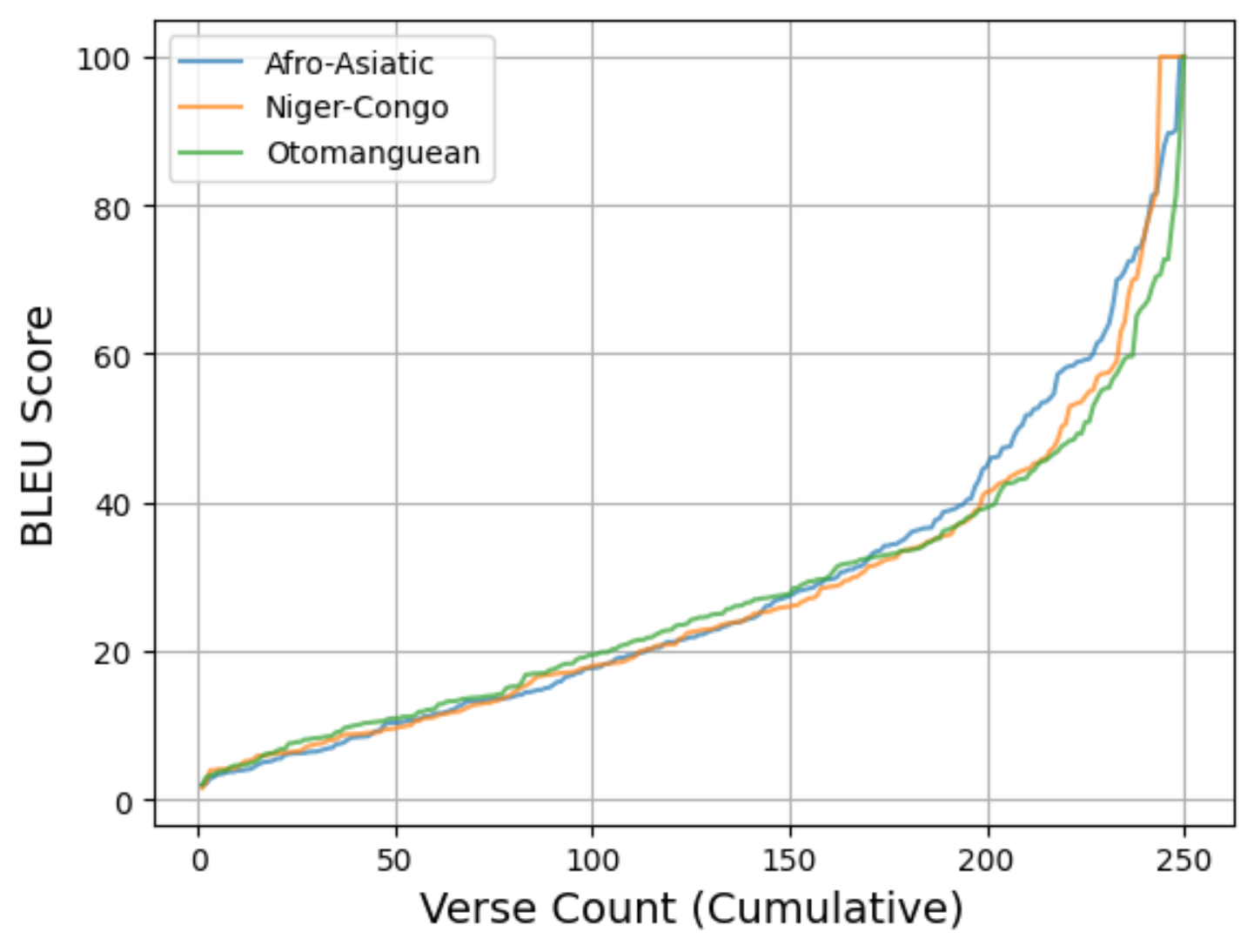}
    \caption{Cumulative per-verse BLEU score distribution for Afro-Asiatic, Niger-Congo, Otomanguean models on \emph{\ac{CV}} task.}
    \label{fig:bleu_breakdown}
\end{figure}

Evaluating median verse accuracy across a range of word, subword and character-level metrics can help to establish a broad intuition about the usefulness of a translation model, but it is also important to evaluate the distribution of these metrics.
Generally, the distribution can be quite broad, as shown in Figure \ref{fig:bleu_breakdown} for the Afro-Asiatic, Niger-Congo, and Otomanguean \emph{\ac{CV}} models.
When the same or similar verse text occurs in multiple passages (e.g., the parables in the Gospels), or when the verse text follows a repeating pattern (“from the tribe of Joseph, ...”; “from the tribe of Dan, ...”), accuracy can be relatively higher.
Translations for longer, more complex verses tend to be relatively lower.
In the context of \ac{BT}, presenting the model’s confidence level to the translator may be as helpful as presenting the suggested verse text, helping the translator know where to focus.
Augmenting the model’s predictions with external evaluation metrics may also be helpful for focusing the translator’s attention on low-confidence verse drafts.
Empirically collected data on challenging verses could also provide a valuable means of focusing the translator’s efforts.

\section{Future Work}
We envision a number of opportunities to expand and improve on this current work.
From a linguistic point of view, our results open up questions about how spBLEU scoring might capture agglutination and scripting more effectively than a BLEU score, as well as broader questions about how the viability of scoring methods might change based on language morphology.
Furthermore, it is not usually clear why certain language families seem more amenable to fine-tuning than others using \ac{NLLB} (as measured by BLEU scores).
It is also unclear whether larger \ac{NLLB} models (such as the next largest architecture with 3.3B tunable parameters) will lead to continued improvements, or yield diminishing returns.

Our work also highlights potential benefits from incorporating multiple \ac{BT}s during model training, although the mixed results indicate that more sophisticated strategies should be investigated, while considering factors such as language and translation characteristics (e.g. translation age, reading level, style) and other metrics (e.g. word alignment scores, subword evenness).
Automated methods for selecting the best source and related language translations may improve performance over the heuristics used in this work.

In this work, changes to the \ac{NLLB} tokenizer were limited to the introduction of special tokens for new language codes.
Some of the translations, such as the Yopno (yut-yut) and Zapotec Tabaa (zat-zatNTps) translations, included a number of characters that were not known to the \ac{NLLB} tokenizer, slightly reducing the accuracy of the fine-tuned model.
Other translations, such as the Limbu translation in Limbu script (lif-lifNT) included many unknown characters, and prevented their use in this work.
We would like to explore translation from the original texts in ancient Greek, Hebrew, and Aramaic, which requires careful attention to a variety of issues involving character encoding, segmentation, and other issues.
Effective methods for augmenting NLLB and the NLLB tokenizer for new scripts and characters will be necessary to achieve the full benefits of the eBible corpus.
We would also like to explore using this parallel corpus to infer characteristics of translation languages for the purpose of language documentation.

Moreover, given the limitations of available data and compute resources for target communities there is a gap in developing techniques that overcome the "low-resource double bind" \cite{ahia2021low}.
Towards this end, implementing and studying efficient fine-tuning methods \cite{pfeiffer2020AdapterHub} would provide significant support for the more advanced neural models to be useful for translators on the ground.

The wide distribution of per-verse translation accuracy indicates the need for effective human-in-the-loop strategies to support the Bible translator, ensuring that these differences are accurately and intuitively presented to guide their work.

\section{Conclusions}
In this work we present the eBible corpus: an open-source \ac{NMT}-ready dataset of over a thousand partial and full \ac{BT}s spanning more than 800 languages.
With the aim of leveraging machine translation to aid expert Bible translators, we introduce a number of benchmark tasks based on the translation ordering often used by these experts.
These tasks include a randomized \ac{CV} along with \ac{NT}- and \ac{OT}-specific objectives.
In addition to benchmark tasks, we provide benchmark model results for selected language families, with models ranging from \ac{SMT} methods to a fine-tuned \ac{NLLB} architecture.

\section{Limitations} 

\section*{Ethics Statement}

\section*{Contributions}
J.M. cast the vision and along with M.M. organized the work on the eBible corpus within the Partnership for Applied Biblical \ac{NLP}.
V.A. wrote the initial scripts to download the corpus from eBible.org while D.B. improved upon these.
U.H. provided significant support for pre-processing through manual and automatic data checks and cleanup.
M.M, J.M, T.J, D.B and M.S. designed the benchmark experiments.
M.M., D.B and D.D. ran experiments and reported results.
M.S. M.M., U.H. and J.M. wrote and edited the manuscript.

\section*{Acknowledgements}
The authors would like to thank their respective organizations for encouraging their participation in this cross-organization collaboration.
They would also like to acknowledge the foundational and ongoing work of Michael Johnson in creating and maintaining the eBible.org site.
Finally, we thank the ETEN Innovation Lab (EIL) for sponsoring the computing resources used in this project.

\bibliography{anthology,custom}

\begin{thebibliography}{25}
\expandafter\ifx\csname natexlab\endcsname\relax\def\natexlab#1{#1}\fi

\bibitem[{Ahia et~al.(2021)Ahia, Kreutzer, and Hooker}]{ahia2021low}
Orevaoghene Ahia, Julia Kreutzer, and Sara Hooker. 2021.
\newblock The low-resource double bind: An empirical study of pruning for
  low-resource machine translation.
\newblock \emph{arXiv preprint arXiv:2110.03036}.

\bibitem[{Akera et~al.(2022)Akera, Mukiibi, Sanyu~Naggayi, Babirye, Owomugisha,
  Nsumba, Nakatumba-Nabende, Bainomugisha, Mwebaze, and
  Quinn}]{akera2022machine}
Benjamin Akera, Jonathan Mukiibi, Lydia Sanyu~Naggayi, Claire Babirye, Isaac
  Owomugisha, Solomon Nsumba, Joyce Nakatumba-Nabende, Engineer Bainomugisha,
  Ernest Mwebaze, and John Quinn. 2022.
\newblock Machine translation for {A}frican languages: Community creation of
  datasets and models in {U}ganda.

\bibitem[{Baines et~al.(2023{\natexlab{a}})Baines, Mathew, and
  Martin}]{ebible_data}
David Baines, Joel Mathew, and Michael Martin. 2023{\natexlab{a}}.
\newblock The e{B}ible corpus.
\newblock \url{https://github.com/BibleNLP/ebible}.

\bibitem[{Baines et~al.(2023{\natexlab{b}})Baines, Mathew, and
  Martin}]{ebible_experiments}
David Baines, Joel Mathew, and Michael Martin. 2023{\natexlab{b}}.
\newblock e{B}ible corpus {NMT} experiments.
\newblock \url{https://github.com/BibleNLP/ebible-experiments}.

\bibitem[{Christodouloupoulos and
  Steedman(2015)}]{christodouloupoulos2015massively}
Christos Christodouloupoulos and Mark Steedman. 2015.
\newblock A massively parallel corpus: the {B}ible in 100 languages.
\newblock \emph{Language resources and evaluation}, 49:375--395.

\bibitem[{Collin(2010)}]{collin2010ethnologue}
Richard~Oliver Collin. 2010.
\newblock Ethnologue.
\newblock \emph{Ethnopolitics}, 9(3-4):425--432.

\bibitem[{Costa-juss{\`a} et~al.(2022)Costa-juss{\`a}, Cross, {\c{C}}elebi,
  Elbayad, Heafield, Heffernan, Kalbassi, Lam, Licht, Maillard
  et~al.}]{costa2022no}
Marta~R Costa-juss{\`a}, James Cross, Onur {\c{C}}elebi, Maha Elbayad, Kenneth
  Heafield, Kevin Heffernan, Elahe Kalbassi, Janice Lam, Daniel Licht, Jean
  Maillard, et~al. 2022.
\newblock No language left behind: Scaling human-centered machine translation.
\newblock \emph{arXiv preprint arXiv:2207.04672}.

\bibitem[{Daspit(2023)}]{silnlp}
Damien Daspit. 2023.
\newblock {SIL-NLP}.
\newblock \url{https://github.com/sillsdev/silnlp}.

\bibitem[{Deng and Byrne(2008)}]{deng2008hmm}
Yonggang Deng and William Byrne. 2008.
\newblock {HMM} word and phrase alignment for statistical machine translation.
\newblock \emph{IEEE Transactions on Audio, Speech, and Language Processing},
  16(3):494--507.

\bibitem[{Dyer et~al.(2013)Dyer, Chahuneau, and Smith}]{dyer2013simple}
Chris Dyer, Victor Chahuneau, and Noah~A Smith. 2013.
\newblock A simple, fast, and effective reparameterization of {IBM} model 2.
\newblock In \emph{Proceedings of the 2013 Conference of the North American
  Chapter of the Association for Computational Linguistics: Human Language
  Technologies}, pages 644--648.

\bibitem[{Goyal et~al.(2022)Goyal, Gao, Chaudhary, Chen, Wenzek, Ju, Krishnan,
  Ranzato, Guzm{\'a}n, and Fan}]{goyal2022flores}
Naman Goyal, Cynthia Gao, Vishrav Chaudhary, Peng-Jen Chen, Guillaume Wenzek,
  Da~Ju, Sanjana Krishnan, Marc’Aurelio Ranzato, Francisco Guzm{\'a}n, and
  Angela Fan. 2022.
\newblock The flores-101 evaluation benchmark for low-resource and multilingual
  machine translation.
\newblock \emph{Transactions of the Association for Computational Linguistics},
  10:522--538.

\bibitem[{Hermjakob(2023)}]{wildebeest}
Ulf Hermjakob. 2023.
\newblock Wildebeest.
\newblock \url{https://github.com/uhermjakob/wildebeest}.

\bibitem[{Kann et~al.(2022)Kann, Ebrahimi, Mager, Oncevay, Ortega, Rios, Fan,
  Gutierrez-Vasques, Chiruzzo, Gim{\'e}nez-Lugo et~al.}]{kann2022americasnli}
Katharina Kann, Abteen Ebrahimi, Manuel Mager, Arturo Oncevay, John~E Ortega,
  Annette Rios, Angela Fan, Ximena Gutierrez-Vasques, Luis Chiruzzo, Gustavo~A
  Gim{\'e}nez-Lugo, et~al. 2022.
\newblock Americasnli: Machine translation and natural language inference
  systems for {I}ndigenous languages of the {A}mericas.
\newblock \emph{Frontiers in Artificial Intelligence}, 5:266.

\bibitem[{Klassen(2022)}]{usfm}
Jeff Klassen. 2022.
\newblock Unified standard format markers.
\newblock \url{https://github.com/ubsicap/usfm}.

\bibitem[{Klein et~al.(2017)Klein, Kim, Deng, Senellart, and
  Rush}]{klein2017opennmt}
Guillaume Klein, Yoon Kim, Yuntian Deng, Jean Senellart, and Alexander~M Rush.
  2017.
\newblock Opennmt: Open-source toolkit for neural machine translation.
\newblock \emph{arXiv preprint arXiv:1701.02810}.

\bibitem[{Koehn(2009)}]{koehn2009statistical}
Philipp Koehn. 2009.
\newblock \emph{Statistical machine translation}.
\newblock Cambridge University Press.

\bibitem[{Liedes(2018)}]{liedes_fairseq}
Sami Liedes. 2018.
\newblock Machine translating the {B}ible into new languages.
\newblock \url{https://github.com/sliedes/fairseq-py}.

\bibitem[{McCarthy et~al.(2020)McCarthy, Wicks, Lewis, Mueller, Wu, Adams,
  Nicolai, Post, and Yarowsky}]{mccarthy2020johns}
Arya~D McCarthy, Rachel Wicks, Dylan Lewis, Aaron Mueller, Winston Wu, Oliver
  Adams, Garrett Nicolai, Matt Post, and David Yarowsky. 2020.
\newblock The {J}ohns {H}opkins {U}niversity {B}ible corpus: 1600+ tongues for
  typological exploration.
\newblock In \emph{Proceedings of the Twelfth Language Resources and Evaluation
  Conference}, pages 2884--2892.

\bibitem[{Ott et~al.(2019)Ott, Edunov, Baevski, Fan, Gross, Ng, Grangier, and
  Auli}]{ott2019fairseq}
Myle Ott, Sergey Edunov, Alexei Baevski, Angela Fan, Sam Gross, Nathan Ng,
  David Grangier, and Michael Auli. 2019.
\newblock fairseq: A fast, extensible toolkit for sequence modeling.
\newblock \emph{arXiv preprint arXiv:1904.01038}.

\bibitem[{Papineni et~al.(2002)Papineni, Roukos, Ward, and
  Zhu}]{papineni2002bleu}
Kishore Papineni, Salim Roukos, Todd Ward, and Wei-Jing Zhu. 2002.
\newblock Bleu: a method for automatic evaluation of machine translation.
\newblock In \emph{Proceedings of the 40th annual meeting of the Association
  for Computational Linguistics}, pages 311--318.

\bibitem[{Pfeiffer et~al.(2020)Pfeiffer, R{\"u}ckl{\'e}, Poth, Kamath,
  Vuli{\'c}, Ruder, Cho, and Gurevych}]{pfeiffer2020AdapterHub}
Jonas Pfeiffer, Andreas R{\"u}ckl{\'e}, Clifton Poth, Aishwarya Kamath, Ivan
  Vuli{\'c}, Sebastian Ruder, Kyunghyun Cho, and Iryna Gurevych. 2020.
\newblock Adapterhub: A framework for adapting transformers.
\newblock In \emph{Proceedings of the 2020 Conference on Empirical Methods in
  Natural Language Processing: System Demonstrations}, pages 46--54.

\bibitem[{Popovi{\'c}(2015)}]{popovic2015chrf}
Maja Popovi{\'c}. 2015.
\newblock chrf: character n-gram {F}-score for automatic {MT} evaluation.
\newblock In \emph{Proceedings of the tenth workshop on statistical machine
  translation}, pages 392--395.

\bibitem[{Resnik et~al.(1999)Resnik, Olsen, and Diab}]{resnik1999bible}
Philip Resnik, Mari~Broman Olsen, and Mona Diab. 1999.
\newblock The {B}ible as a parallel corpus: Annotating the ‘{B}ook of 2000
  {T}ongues’.
\newblock \emph{Computers and the Humanities}, 33:129--153.

\bibitem[{Wu et~al.(2018)Wu, Vyas, and Yarowsky}]{wu2018creating}
Winston Wu, Nidhi Vyas, and David Yarowsky. 2018.
\newblock Creating a translation matrix of the {B}ible’s names across 591
  languages.
\newblock In \emph{Proceedings of the Eleventh International Conference on
  Language Resources and Evaluation (LREC 2018)}.

\bibitem[{Zhu et~al.(2020)Zhu, Xia, Wu, He, Qin, Zhou, Li, and
  Liu}]{zhu2020incorporating}
Jinhua Zhu, Yingce Xia, Lijun Wu, Di~He, Tao Qin, Wengang Zhou, Houqiang Li,
  and Tie-Yan Liu. 2020.
\newblock Incorporating bert into neural machine translation.
\newblock \emph{arXiv preprint arXiv:2002.06823}.

\end{thebibliography}
\bibliographystyle{acl_natbib}

\appendix

\section{Characteristics of Selected Bible Translations}
\label{sec:BT_chars}
Table \ref{tab:bible_translation_characteristics} goes into greater depth on the characteristics of selected \ac{BT}s.
\begin{table*}[]
\centering
\resizebox{\textwidth}{!}{%
\begin{tabular}{|c|c|c|c|c|c|c|c|c|c|}
\hline
\textbf{\begin{tabular}[c]{@{}c@{}}Language \\ Family\end{tabular}} &
  \textbf{Purpose} &
  \textbf{Language} &
  \textbf{\begin{tabular}[c]{@{}c@{}}ISO- \\ 639-3\end{tabular}} &
  \textbf{\begin{tabular}[c]{@{}c@{}}eBible\\ Translation\end{tabular}} &
  \textbf{Scope} &
  \textbf{Verses} &
  \textbf{Script} &
  \textbf{Typology} &
  \textbf{Country} \\ \hline
\multirow{3}{*}{Afro-Asiatic} &
  Source &
  Hausa &
  hau &
  hau-hausa.txt &
  Bible &
  31,082 &
  Latn &
  SVO &
  Nigeria \\ \cline{2-10} 
 &
  Target &
  Dangaléat &
  daa &
  daa-daaNT.txt &
  NT &
  7,957 &
  Latn &
  Unk &
  Chad \\ \cline{2-10} 
 &
  Related &
  \begin{tabular}[c]{@{}c@{}}Fulfulde,\\ Western Niger\end{tabular} &
  fuh &
  fuh-fuhbkf.txt &
  NT &
  7,57 &
  Latn &
  SVO &
  Niger \\ \hline
\multirow{3}{*}{Austronesian} &
  Source &
  Kuanua &
  ksd &
  ksd-ksd.txt &
  Bible &
  31,098 &
  Latn &
  Unk &
  PNG* \\ \cline{2-10} 
 &
  Target &
  Kandas &
  kqw &
  kqw-kqw.txt &
  NT &
  7,957 &
  Latn &
  Unk &
  PNG \\ \cline{2-10} 
 &
  Related &
  Ramoaaina &
  rai &
  rai-rai.txt &
  NT &
  7,957 &
  Latn &
  SVO &
  PNG \\ \hline
\multirow{3}{*}{Dravidian} &
  Source &
  Tamil &
  tam &
  tam-tam2017.txt &
  Bible &
  31,099 &
  Taml &
  SOV &
  India \\ \cline{2-10} 
 &
  Target &
  Malayalam &
  mal &
  mal-mal.txt &
  Bible &
  31,089 &
  Mlym &
  SOV &
  India \\ \cline{2-10} 
 &
  Related &
  Kannada &
  kan &
  kan-kan2017.txt &
  Bible &
  31,099 &
  Knda &
  SOV &
  India \\ \hline
\multirow{3}{*}{Indo-European} &
  Source &
  Hindi &
  hin &
  hin-hin2017.txt &
  Bible &
  31,099 &
  Deva &
  SOV &
  India \\ \cline{2-10} 
 &
  Target &
  \begin{tabular}[c]{@{}c@{}}Eastern\\ Panjabi\end{tabular} &
  pan &
  pan-pan.txt &
  Bible &
  31,099 &
  Gurm &
  SOV &
  India \\ \cline{2-10} 
 &
  Related &
  Gujarati &
  guj &
  guj-guj2017.txt &
  Bible &
  31,099 &
  Gujr &
  SOV &
  India \\ \hline
\multirow{3}{*}{Niger-Congo} &
  Source &
  Swahili &
  swh &
  swh-swhonen.txt &
  Bible &
  31,098 &
  Latn &
  SVO &
  Tanzania \\ \cline{2-10} 
 &
  Target &
  Kwere &
  cwe &
  cwe-cwe.txt &
  \begin{tabular}[c]{@{}c@{}}GEN-DEU, \\ NT\end{tabular} &
  13,806 &
  Latn &
  Unk &
  Tanzania \\ \cline{2-10} 
 &
  Related &
  Vidunda &
  vid &
  vid-vid.txt &
  \begin{tabular}[c]{@{}c@{}}GEN-DEU, \\ NT\end{tabular} &
  13,809 &
  Latn &
  Unk &
  Tanzania \\ \hline
\multirow{3}{*}{Otomanguean} &
  Source &
  Spanish &
  spa &
  spa-sparvg.txt &
  Bible &
  31,097 &
  Latn &
  SVO &
  Spain \\ \cline{2-10} 
 &
  Target &
  \begin{tabular}[c]{@{}c@{}}Zapotec,\\ Tabaa\end{tabular} &
  zat &
  zat-zatNTps.txt &
  PSA, NT &
  10,416 &
  Latn &
  VSO &
  Mexico \\ \cline{2-10} 
 &
  Related &
  \begin{tabular}[c]{@{}c@{}}Tapotec,\\ Cajonos\end{tabular} &
  zad &
  zad-zadNT.txt &
  NT &
  7,957 &
  Latn &
  VSO &
  Mexico \\ \hline
\multirow{3}{*}{Sino-Tibetan} &
  Source &
  Nepali &
  npi &
  npi-npiulb.txt &
  Bible &
  31,099 &
  Deva &
  SOV &
  Nepal \\ \cline{2-10} 
 &
  Target &
  \begin{tabular}[c]{@{}c@{}}Tamang,\\ Eastern\end{tabular} &
  taj &
  taj-taj.txt &
  NT &
  7,957 &
  Deva &
  SOV &
  Nepal \\ \cline{2-10} 
 &
  Related &
  Limbu &
  lif &
  lif-lifNT2.txt &
  NT &
  7,957 &
  Deva &
  SOV &
  Nepal \\ \hline
\multirow{3}{*}{Trans-New Guinea} &
  Source &
  Tok Pisin &
  tpi &
  tpi-tpiOTNT.txt &
  Bible &
  31,099 &
  Latn &
  SOV &
  PNG \\ \cline{2-10} 
 &
  Target &
  Yopno &
  yut &
  yut-yut.txt &
  PSA, NT &
  10,417 &
  Latn &
  SOV &
  PNG \\ \cline{2-10} 
 &
  Related &
  Iyo &
  nca &
  nca-nca.txt &
  NT &
  7,957 &
  Latn &
  SOV &
  PNG \\ \hline
\end{tabular}%
}
\caption{\label{tab:bible_translation_characteristics}
\Ac{BT} characteristics.
}
\end{table*}

\section{NLLB Language Codes}
\label{sec:NLLB_lang_codes}
Table \ref{tab:nllb_lang_codes} lists the language codes specific to Meta's \ac{NLLB}.

\begin{table*}[]
\centering
\resizebox{\textwidth}{!}{%
\begin{tabular}{|c|c|c|c|}
\hline
\textbf{Language Family} & \textbf{Language} & \textbf{ISO-639-3} & \textbf{\begin{tabular}[c]{@{}c@{}}NLLB \\ Language Code\end{tabular}} \\ \hline
\multirow{3}{*}{Afro-Asiatic}     & Hausa                   & hau & hau\_Latn     \\ \cline{2-4} 
                                  & Dangaléat               & daa & (!) daa\_Latn \\ \cline{2-4} 
                                  & Fulfulde, Western Niger & fuh & (!) fuh\_Latn \\ \hline
\multirow{3}{*}{Austronesian}     & Kuanua                  & ksd & (!) ksd\_Latn \\ \cline{2-4} 
                                  & Kandas                  & kqw & (!) kqw\_Latn \\ \cline{2-4} 
                                  & Ramoaaina               & rai & (!) rai\_Latn \\ \hline
\multirow{3}{*}{Dravidian}        & Tamil                   & tam & tam\_Taml     \\ \cline{2-4} 
                                  & Malayalam               & mal & mal\_Mlym     \\ \cline{2-4} 
                                  & Kannada                 & kan & kan\_Knda     \\ \hline
\multirow{3}{*}{Indo-European}    & Hindi                   & hin & hin\_Deva     \\ \cline{2-4} 
                                  & Eastern Panjabi         & pan & pan\_Gurm     \\ \cline{2-4} 
                                  & Gujarati                & guj & guj\_Gujr     \\ \hline
\multirow{3}{*}{Niger-Congo}      & Swahili                 & swh & swh\_Latn     \\ \cline{2-4} 
                                  & Kwere                   & cwe & (!) cwe\_Latn \\ \cline{2-4} 
                                  & Vidunda                 & vid & (!) vid\_Latn \\ \hline
\multirow{3}{*}{Otomanguean}      & Spanish                 & spa & spa\_Latn     \\ \cline{2-4} 
                                  & Zapotec, Tabaa          & zat & (!) zat\_Latn \\ \cline{2-4} 
                                  & Tapotec, Cajonos        & zad & (!) zad\_Latn \\ \hline
\multirow{3}{*}{Sino-Tibetan}     & Nepali                  & npi & npi\_Deva     \\ \cline{2-4} 
                                  & Tamang, Eastern         & taj & (!) taj\_Deva \\ \cline{2-4} 
                                  & Limbu                   & lif & (!) lif\_Deva \\ \hline
\multirow{3}{*}{Trans-New Guinea} & Tok Pisin               & tpi & tpi\_Latn     \\ \cline{2-4} 
                                  & Yopno                   & yut & (!) yut\_Latn \\ \cline{2-4} 
                                  & Iyo                     & nca & (!) nca\_Latn \\ \hline
\end{tabular}%
}
\caption{\label{tab:nllb_lang_codes}
\Ac{NLLB} language codes.
}
\end{table*}

\section{Source / Target / Related Language Alignment Scores (\Ac{HMM})}
We report on initial \ac{HMM} scores which led to our decisions around specific source and target language decisions within selected language families.

\subsection{Afro-Asiatic}
For the Afro-Asiatic language family, the \emph{Chadic > Biu-Mandara} branch contains 79 languages (four translations in the corpus) and the \emph{Chadic > East} branch contains 36 languages (one translation in the corpus).
These five translations only contain the \ac{NT} portion of the Bible, and are languages spoken in either Cameroon or Chad.
English and French are national languages for Cameroon. French and Arabic are national languages for Chad.
Table \ref{tab:hmm_afro_asiatic} shows that the best alignment results were achieved using a Hausa translation (hau-hausa) with the Dangaléat translation (daa-daaNT).
The Western Niger Fulfulde translation (ful-fuhbkf) aligned best with the Dangaléat translation.

\begin{table}[]
\centering
\resizebox{0.49\textwidth}{!}{%
\begin{tabular}{|lc|rrrr|rr|}
\hline
\multicolumn{2}{|c|}{\textbf{\begin{tabular}[c]{@{}c@{}}Target\\ Language(s)\end{tabular}}} &
  \multicolumn{4}{c|}{\textbf{\begin{tabular}[c]{@{}c@{}}National/Gateway\\ Language(s)\end{tabular}}} &
  \multicolumn{2}{c|}{\textbf{\begin{tabular}[c]{@{}c@{}}Related\\ Language(s)\end{tabular}}} \\ \hline
\multicolumn{1}{|c|}{\textbf{Language}} &
  \textbf{Translation} &
  \multicolumn{1}{c|}{\textbf{eng-engULB}} &
  \multicolumn{1}{c|}{\textbf{fra-frasbl}} &
  \multicolumn{1}{c|}{\textbf{hau-hausa}} &
  \multicolumn{1}{c|}{\textbf{hau-hauulb}} &
  \multicolumn{1}{c|}{\textbf{ffm-ffm}} &
  \multicolumn{1}{c|}{\textbf{fuh-fhubkf}} \\ \hline
\multicolumn{1}{|l|}{Hdi} &
  xed-xed &
  \multicolumn{1}{r|}{0.1792} &
  \multicolumn{1}{r|}{0.1691} &
  \multicolumn{1}{r|}{0.1860} &
  0.1736 &
  \multicolumn{1}{r|}{0.1915} &
  0.1982 \\ \hline
\multicolumn{1}{|l|}{Mbuko} &
  mqb-mqbNT &
  \multicolumn{1}{r|}{0.1366} &
  \multicolumn{1}{r|}{0.1291} &
  \multicolumn{1}{r|}{0.1465} &
  0.1298 &
  \multicolumn{1}{r|}{0.1560} &
  0.1613 \\ \hline
\multicolumn{1}{|l|}{Merey} &
  meq-meq &
  \multicolumn{1}{r|}{0.1422} &
  \multicolumn{1}{r|}{0.1368} &
  \multicolumn{1}{r|}{0.1546} &
  0.1376 &
  \multicolumn{1}{r|}{0.1598} &
  0.1640 \\ \hline
\multicolumn{1}{|l|}{Muyang} &
  muy-muy &
  \multicolumn{1}{r|}{0.1513} &
  \multicolumn{1}{r|}{0.1451} &
  \multicolumn{1}{r|}{0.1537} &
  0.1403 &
  \multicolumn{1}{r|}{0.1818} &
  0.1862 \\ \hline
\multicolumn{1}{|l|}{Dangaléat} &
  daa-daaNT &
  \multicolumn{1}{r|}{0.1912} &
  \multicolumn{1}{r|}{0.1760} &
  \multicolumn{1}{r|}{\textbf{0.1929}} &
  0.1705 &
  \multicolumn{1}{r|}{0.2044} &
  \textbf{0.2148} \\ \hline
\end{tabular}%
}
\caption{\label{tab:hmm_afro_asiatic}
Source / Target / Related Language Alignment Scores (Afro-Asiatic).
}
\end{table}

\subsection{Austronesian}
For the Austronesian language family, the \emph{Malayo-Polynesian > Central-Eastern Malayo-Polynesian > Eastern Malayo-Polynesian > Oceanic} branch contains 513 languages.
Of these 513 languages, the eBible corpus contains four translations from the Western Oceanic sub-branch.
All four of these are for languages spoken in Papua New Guinea (PNG).
Three of these translations contain the \ac{NT} only, and one contains the \ac{NT} and a portion of the \ac{OT} (four translations in the corpus) and the \emph{Chadic > East} branch contains 36 languages (one translation in the corpus).
English and Tok Pisin (tpi) are national languages in PNG, while Dobu (dob), Kuanua (ksd), Suau (swp) and Tawala (tbo) are gateway languages from the \emph{Malayo-Polynesian} branch with translations in the corpus.

Among these translations, the Kuanua translation (ksd-ksd) was selected as the source and the Kandas translation (kqw-kqw) was selected as the target due to their strong alignment score.
The Label (lbb-lbb) and Ramoaaina (rai-rai) translations both aligned well with the Kandas translation (kqw-kqw); Ramoaaina was selected as the related language translation.
Results are summarized in Table \ref{tab:hmm_austronesian}.

\begin{table}[]
\centering
\resizebox{0.49\textwidth}{!}{%
\begin{tabular}{|lc|rrrrr|lllrr|}
\hline
\multicolumn{2}{|c|}{\textbf{\begin{tabular}[c]{@{}c@{}}Target\\ Language(s)\end{tabular}}} &
  \multicolumn{5}{c|}{\textbf{\begin{tabular}[c]{@{}c@{}}National/Gateway\\ Language(s)\end{tabular}}} &
  \multicolumn{5}{c|}{\textbf{\begin{tabular}[c]{@{}c@{}}Related\\ Language(s)\end{tabular}}} \\ \hline
\multicolumn{1}{|c|}{\textbf{Language}} &
  \textbf{Translation} &
  \multicolumn{1}{c|}{\textbf{tpi-tpiOTNT}} &
  \multicolumn{1}{c|}{\textbf{dob-dob}} &
  \multicolumn{1}{c|}{\textbf{swp-swp}} &
  \multicolumn{1}{c|}{\textbf{tbo-tbo}} &
  \multicolumn{1}{c|}{\textbf{ksd-ksd}} &
  \multicolumn{1}{c|}{\textbf{kqw-kqw}} &
  \multicolumn{1}{c|}{\textbf{lbb-lbb}} &
  \multicolumn{1}{c|}{\textbf{gfk-gfk}} &
  \multicolumn{1}{c|}{\textbf{rai-rai}} &
  \multicolumn{1}{c|}{\textbf{sgq-sgq}} \\ \hline
\multicolumn{1}{|l|}{Fanamarket} &
  bjp-bjp &
  \multicolumn{1}{r|}{0.2581} &
  \multicolumn{1}{r|}{0.1761} &
  \multicolumn{1}{r|}{0.2068} &
  \multicolumn{1}{r|}{0.1990} &
  0.2701 &
  \multicolumn{1}{r|}{0.3994} &
  \multicolumn{1}{r|}{0.4134} &
  \multicolumn{1}{r|}{0.3025} &
  \multicolumn{1}{r|}{0.3974} &
  0.2576 \\ \hline
\multicolumn{1}{|l|}{Kandas} &
  kqw-kqw &
  \multicolumn{1}{r|}{0.2213} &
  \multicolumn{1}{r|}{0.1629} &
  \multicolumn{1}{r|}{0.2003} &
  \multicolumn{1}{r|}{0.1897} &
  \textbf{0.3303} &
  \multicolumn{1}{l|}{N/A} &
  \multicolumn{1}{r|}{\textbf{0.6847}} &
  \multicolumn{1}{r|}{0.2910} &
  \multicolumn{1}{r|}{\textbf{0.6849}} &
  0.2516 \\ \hline
\multicolumn{1}{|l|}{Label} &
  lbb-lbb &
  \multicolumn{1}{r|}{0.2152} &
  \multicolumn{1}{r|}{0.1657} &
  \multicolumn{1}{r|}{0.2127} &
  \multicolumn{1}{r|}{0.1979} &
  0.3229 &
  \multicolumn{1}{l|}{N/A} &
  \multicolumn{1}{l|}{N/A} &
  \multicolumn{1}{r|}{0.2789} &
  \multicolumn{1}{r|}{0.6583} &
  0.2538 \\ \hline
\multicolumn{1}{|l|}{Patpatar} &
  gfk-gfk &
  \multicolumn{1}{r|}{0.1966} &
  \multicolumn{1}{r|}{0.1462} &
  \multicolumn{1}{r|}{01662} &
  \multicolumn{1}{r|}{0.1567} &
  0.2707 &
  \multicolumn{1}{l|}{N/A} &
  \multicolumn{1}{l|}{N/A} &
  \multicolumn{1}{l|}{N/A} &
  \multicolumn{1}{r|}{0.2938} &
  0.1919 \\ \hline
\multicolumn{1}{|l|}{Ramoaaina} &
  rai-rai &
  \multicolumn{1}{r|}{0.2120} &
  \multicolumn{1}{r|}{0.1613} &
  \multicolumn{1}{r|}{0.2069} &
  \multicolumn{1}{r|}{0.1874} &
  0.3916 &
  \multicolumn{1}{l|}{N/A} &
  \multicolumn{1}{l|}{N/A} &
  \multicolumn{1}{l|}{N/A} &
  \multicolumn{1}{l|}{N/A} &
  0.2475 \\ \hline
\multicolumn{1}{|l|}{Sursurunga} &
  sgq-sgq &
  \multicolumn{1}{r|}{0.1775} &
  \multicolumn{1}{r|}{0.1353} &
  \multicolumn{1}{r|}{0.1440} &
  \multicolumn{1}{r|}{0.1498} &
  0.1692 &
  \multicolumn{1}{l|}{N/A} &
  \multicolumn{1}{l|}{N/A} &
  \multicolumn{1}{l|}{N/A} &
  \multicolumn{1}{l|}{N/A} &
  \multicolumn{1}{l|}{N/A} \\ \hline
\end{tabular}%
}
\caption{\label{tab:hmm_austronesian}
Source / Target / Related Language Alignment Scores (Austronesian).
}
\end{table}

\subsection{Dravidian}
For the Dravidian language family (85 total languages), the \emph{Southern > Tamil-Kannada} branch contains 31 languages.
Of these 31 languages, the eBible corpus contains five translations, one from the Kannada sub-branch and four from the \emph{Tamil-Kodagu} sub-branch.
All five of these are for languages spoken in India, and each translation is a full \ac{BT}.
Each of these translations is for a national language of India (Kannada, Malayalam, Tamil, and Telugu).
There are no translations for low-resource languages from this language family and geography available in the corpus.
As a result, the Tamil (tam-tam2017) and Malayalam (mal-mal) translations were chosen for the source and target translation pairing, with the Kannada translation (kan-kan2017) as the related language translation.
Results are summarized in Table \ref{tab:hmm_dravidian}.

\begin{table}[]
\centering
\resizebox{0.49\textwidth}{!}{%
\begin{tabular}{|lc|c|crr|}
\hline
\multicolumn{2}{|c|}{\textbf{\begin{tabular}[c]{@{}c@{}}Target\\ Language(s)\end{tabular}}} &
  \textbf{\begin{tabular}[c]{@{}c@{}}National/Gateway\\ Language(s)\end{tabular}} &
  \multicolumn{3}{c|}{\textbf{\begin{tabular}[c]{@{}c@{}}Related\\ Language(s)\end{tabular}}} \\ \hline
\multicolumn{1}{|c|}{\textbf{Language}} &
  \textbf{Translation} &
  \textbf{hin-hin2017} &
  \multicolumn{1}{c|}{\textbf{kan-kan2017}} &
  \multicolumn{1}{c|}{\textbf{mal-mal}} &
  \multicolumn{1}{c|}{\textbf{mal-malc}} \\ \hline
\multicolumn{1}{|l|}{Tamil}  & tam-tam2017 & \multicolumn{1}{r|}{0.2063} & \multicolumn{1}{r|}{\textbf{0.3466}} & \multicolumn{1}{r|}{\textbf{0.4295}} & 0.3396 \\ \hline
\multicolumn{1}{|l|}{Telugu} & tel-tel2017 & \multicolumn{1}{r|}{0.2097} & \multicolumn{1}{r|}{0.3423}          & \multicolumn{1}{r|}{0.3229}          & 0.3068 \\ \hline
\end{tabular}%
}
\caption{\label{tab:hmm_dravidian}
Source / Target / Related Language Alignment Scores (Dravidian).
}
\end{table}

\subsection{Indo-European}
For the Indo-European language family, the \emph{Indo-Iranian > Indo-Aryan}  branch contains 220 languages, with 92 languages in the \emph{Intermediate > Western} sub-branch, 94 languages in the \emph{Outer} sub-branch, and 11 languages in the \emph{Western Hindi} sub-branch.
The eBible corpus contains two translations from the \emph{Intermediate > Western} sub-branch, four from the \emph{Outer} sub-branch and five from the Western Hindi sub-branch (representing two languages).
Each of these translations is for a language spoken in Bangladesh, India, Nepal, and/or Pakistan, and each is a full \ac{BT}.
There are no translations for low-resource languages from this language family and geography available in the corpus.
As a result, the Hindi (hin-hin2017) and Eastern Panjabi (pan-pan) translations were chosen for the source and target translation pairing, with the Gujarati translation (guj-gju2017) as the related language translation.
Results are summarized in Table \ref{tab:hmm_indo_european}.

\begin{table}[]
\centering
\resizebox{0.49\textwidth}{!}{%
\begin{tabular}{|ll|rr|rr|}
\hline
\multicolumn{2}{|c|}{\textbf{\begin{tabular}[c]{@{}c@{}}Target\\ Language(s)\end{tabular}}} &
  \multicolumn{2}{c|}{\textbf{\begin{tabular}[c]{@{}c@{}}National/Gateway\\ Language(s)\end{tabular}}} &
  \multicolumn{2}{c|}{\textbf{\begin{tabular}[c]{@{}c@{}}Related\\ Language(s)\end{tabular}}} \\ \hline
\multicolumn{1}{|c|}{\textbf{Language}} &
  \multicolumn{1}{c|}{\textbf{Translation}} &
  \multicolumn{1}{c|}{\textbf{hin-hin2017}} &
  \multicolumn{1}{c|}{\textbf{npi-npiulb}} &
  \multicolumn{1}{c|}{\textbf{pan-pan}} &
  \multicolumn{1}{c|}{\textbf{ben-ben2017}} \\ \hline
\multicolumn{1}{|l|}{Gujarati} &
  guj-guj2017 &
  \multicolumn{1}{r|}{0.3012} &
  0.3200 &
  \multicolumn{1}{r|}{\textbf{0.2973}} &
  0.3041 \\ \hline
\multicolumn{1}{|l|}{E. Panjabi} &
  pan-pan &
  \multicolumn{1}{r|}{\textbf{0.4240}} &
  0.2623 &
  \multicolumn{1}{l|}{N/A} &
  0.3665 \\ \hline
\multicolumn{1}{|l|}{Assamese} &
  asm-asmfb &
  \multicolumn{1}{r|}{0.2796} &
  0.3172 &
  \multicolumn{1}{r|}{0.2616} &
  0.3666 \\ \hline
\multicolumn{1}{|l|}{Bengali} &
  ben-ben2017 &
  \multicolumn{1}{r|}{0.2904} &
  0.3242 &
  \multicolumn{1}{r|}{0.2774} &
  \multicolumn{1}{l|}{N/A} \\ \hline
\multicolumn{1}{|l|}{Marathi} &
  mar-mar &
  \multicolumn{1}{r|}{0.2521} &
  0.2989 &
  \multicolumn{1}{r|}{0.2461} &
  0.2767 \\ \hline
\multicolumn{1}{|l|}{Orya} &
  ory-ory &
  \multicolumn{1}{r|}{0.3036} &
  0.3098 &
  \multicolumn{1}{r|}{0.2942} &
  0.4031 \\ \hline
\multicolumn{1}{|l|}{Urdu} &
  urd-urd &
  \multicolumn{1}{r|}{0.4622} &
  0.2576 &
  \multicolumn{1}{r|}{0.4415} &
  0.2765 \\ \hline
\multicolumn{1}{|l|}{Urdu} &
  urd-urdgvh &
  \multicolumn{1}{r|}{0.3123} &
  0.2346 &
  \multicolumn{1}{r|}{0.2949} &
  0.2270 \\ \hline
\multicolumn{1}{|l|}{Urdu} &
  urd-urdgvr &
  \multicolumn{1}{r|}{0.3092} &
  0.2293 &
  \multicolumn{1}{r|}{0.3012} &
  0.2238 \\ \hline
\multicolumn{1}{|l|}{Urdu} &
  urd-urdgvu &
  \multicolumn{1}{r|}{0.3041} &
  0.2253 &
  \multicolumn{1}{r|}{0.2964} &
  0.2205 \\ \hline
\end{tabular}%
}
\caption{\label{tab:hmm_indo_european}
Source / Target / Related Language Alignment Scores (Indo-European).
}
\end{table}

\subsection{Niger-Congo}
For the Niger-Congo language family, the \emph{Volta-Congo > Benue-Congo > Bantoid > Southern > Narrow-Bantu > Central} branch contains 354 languages.
Of these 354 languages, the eBible corpus contains 19 translations for languages spoken in Tanzania, including three Swahili full \ac{BT}s, two \ac{NT}+ translations (Kwere (cwe-cwe) and Vidunda (vid-vid)), and 14 \ac{NT}-only translations.
Among these translations, the Swahili translation (swh-swhonen) was selected as the source and the Kwere translation (cwe-cwe) was selected as the target, and the Vidunda translation (vid-vid) was selected as the related language translation.
Preference was given to the Kwere and Vidunda translations due to their partial \ac{OT} content.
Results are summarized in Table \ref{tab:hmm_niger_congo}.

\begin{table}[]
\centering
\resizebox{0.49\textwidth}{!}{%
\begin{tabular}{|lc|rr|rr|}
\hline
\multicolumn{2}{|c|}{\textbf{\begin{tabular}[c]{@{}c@{}}Target\\ Language(s)\end{tabular}}} &
  \multicolumn{2}{c|}{\textbf{\begin{tabular}[c]{@{}c@{}}National/Gateway\\ Language(s)\end{tabular}}} &
  \multicolumn{2}{c|}{\textbf{\begin{tabular}[c]{@{}c@{}}Related\\ Language(s)\end{tabular}}} \\ \hline
\multicolumn{1}{|c|}{\textbf{Language}} &
  \textbf{Translation} &
  \multicolumn{1}{c|}{\textbf{swh-swhonen}} &
  \multicolumn{1}{c|}{\textbf{swh-swhulb}} &
  \multicolumn{1}{c|}{\textbf{cwe-cwe}} &
  \multicolumn{1}{c|}{\textbf{vid-vid}} \\ \hline
\multicolumn{1}{|l|}{Kwere*} &
  cwe-cwe &
  \multicolumn{1}{r|}{\textbf{0.4382}} &
  0.3758 &
  \multicolumn{1}{l|}{N/A} &
  \textbf{0.4912} \\ \hline
\multicolumn{1}{|l|}{Isanzu}  & isn-isn & \multicolumn{1}{r|}{0.3245} & 0.5625 & \multicolumn{1}{r|}{0.2638} & 0.2424 \\ \hline
\multicolumn{1}{|l|}{Kutu}    & kdc-kdc & \multicolumn{1}{r|}{0.4326} & 0.3685 & \multicolumn{1}{r|}{0.5347} & 0.4760 \\ \hline
\multicolumn{1}{|l|}{Makonda} & kde-kde & \multicolumn{1}{r|}{0.4509} & 0.3672 & \multicolumn{1}{r|}{0.4451} & 0.4315 \\ \hline
\multicolumn{1}{|l|}{Kisi}    & kiz-kiz & \multicolumn{1}{r|}{0.4320} & 0.7855 & \multicolumn{1}{r|}{0.3446} & 0.3304 \\ \hline
\multicolumn{1}{|l|}{Mwera}   & mwe-mwe & \multicolumn{1}{r|}{0.4216} & 0.3560 & \multicolumn{1}{r|}{0.4287} & 0.4382 \\ \hline
\multicolumn{1}{|l|}{Ndamba}  & ndj-ndj & \multicolumn{1}{r|}{0.4098} & 0.3544 & \multicolumn{1}{r|}{0.4609} & 0.4808 \\ \hline
\multicolumn{1}{|l|}{Ngulu}   & ngp-ngp & \multicolumn{1}{r|}{0.3891} & 0.3383 & \multicolumn{1}{r|}{0.4459} & 0.4112 \\ \hline
\multicolumn{1}{|l|}{Ngindo}  & nnq-nnq & \multicolumn{1}{r|}{0.4118} & 0.3517 & \multicolumn{1}{r|}{0.4602} & 0.4754 \\ \hline
\multicolumn{1}{|l|}{Pogolo}  & poy-poy & \multicolumn{1}{r|}{0.4235} & 0.3608 & \multicolumn{1}{r|}{0.4745} & 0.4781 \\ \hline
\multicolumn{1}{|l|}{Kara}    & reg-reg & \multicolumn{1}{r|}{0.3732} & 0.5998 & \multicolumn{1}{r|}{0.3143} & 0.2850 \\ \hline
\multicolumn{1}{|l|}{Luguru}  & ruf-ruf & \multicolumn{1}{r|}{0.4457} & 0.3807 & \multicolumn{1}{r|}{0.5570} & 0.4701 \\ \hline
\multicolumn{1}{|l|}{Vidunda*} &
  vid-vid &
  \multicolumn{1}{r|}{0.4045} &
  0.3462 &
  \multicolumn{1}{r|}{\textbf{0.4913}} &
  \multicolumn{1}{l|}{N/A} \\ \hline
\multicolumn{1}{|l|}{Vwanji}  & wbi-wbi & \multicolumn{1}{r|}{0.3477} & 0.4811 & \multicolumn{1}{r|}{0.3169} & 0.2933 \\ \hline
\multicolumn{1}{|l|}{Zaramo}  & zaj-zaj & \multicolumn{1}{r|}{0.4652} & 0.3965 & \multicolumn{1}{r|}{0.4784} & 0.4331 \\ \hline
\multicolumn{1}{|l|}{Zigula}  & ziw-ziw & \multicolumn{1}{r|}{0.4149} & 0.3624 & \multicolumn{1}{r|}{0.4748} & 0.4451 \\ \hline
\end{tabular}%
}
\caption{\label{tab:hmm_niger_congo}
Source / Target / Related Language Alignment Scores (Niger-Congo).
}
\end{table}

\subsection{Otomanguean}
For the Otomanguean language family with 179 total languages, the \emph{Eastern-Otomanguean > Popolocan-Zapotecan > Zapotecan} branch contains 64 languages.
Of these 64 languages, the eBible corpus contains 28 translations for languages spoken in Mexico, including three ac{NT}+ translations\footnote{Zapotec Rincón (zar-zarNT), Zapotec Tabaa (zat-zatNTps), and Zapotec Yatee (zty-ztyNTps).}, and 25 \ac{NT}-only translations; there are no full \ac{BT}s from this branch.
Among these translations, a Spanish translation (spa-sparvg) translation was selected as the source and the Zapotec Tabaa translation (zat-zatNTps) was selected as the target; preference was given to the Zaopotec Tabaa translation due to its partial \ac{OT} content.
The Zapotec Cajonos translation (zad-zadNT) was selected as the related language translation.
Results are summarized in Table \ref{tab:hmm_otomanguean}.

\begin{table}[]
\centering
\resizebox{0.49\textwidth}{!}{%
\begin{tabular}{|ll|r|rr|}
\hline
\multicolumn{2}{|c|}{\textbf{\begin{tabular}[c]{@{}c@{}}Target\\ Language(s)\end{tabular}}} &
  \multicolumn{1}{c|}{\textbf{\begin{tabular}[c]{@{}c@{}}National/Gateway\\ Language(s)\end{tabular}}} &
  \multicolumn{2}{c|}{\textbf{\begin{tabular}[c]{@{}c@{}}Related\\ Language(s)\end{tabular}}} \\ \hline
\multicolumn{1}{|c|}{\textbf{Language}} &
  \multicolumn{1}{c|}{\textbf{Translation}} &
  \multicolumn{1}{c|}{\textbf{spa-sparvg}} &
  \multicolumn{1}{c|}{\textbf{zar-zarNT}} &
  \multicolumn{1}{c|}{\textbf{zat-zatNTps}} \\ \hline
\multicolumn{1}{|l|}{Chatino, Tataltepec}               & cta-ctaNT    & 0.1029 & \multicolumn{1}{r|}{0.1229} & 0.1214          \\ \hline
\multicolumn{1}{|l|}{Chatino, Western Highland}         & ctp-ctpNT    & 0.0947 & \multicolumn{1}{r|}{0.1165} & 0.1157          \\ \hline
\multicolumn{1}{|l|}{Chatino, Nopala}                   & cya-cya      & 0.1718 & \multicolumn{1}{r|}{0.1718} & 0.1667          \\ \hline
\multicolumn{1}{|l|}{Zapotec, Sierra de Juárez}         & zaa-zaaNT    & 0.1662 & \multicolumn{1}{r|}{0.1663} & 0.1653          \\ \hline
\multicolumn{1}{|l|}{Zapotec, Western Tlacolula Valley} & zab-zabNT    & 0.2092 & \multicolumn{1}{r|}{0.2281} & 0.2456          \\ \hline
\multicolumn{1}{|l|}{Zapotec, Ocotlán}                  & zac-zacNT    & 0.1733 & \multicolumn{1}{r|}{0.1896} & 0.2022          \\ \hline
\multicolumn{1}{|l|}{Zapotec, Cajonos}                  & zad-zadNT    & 0.1877 & \multicolumn{1}{r|}{0.2485} & \textbf{0.2757} \\ \hline
\multicolumn{1}{|l|}{Zapotec, Isthmus}                  & zai-zaiNT    & 0.1835 & \multicolumn{1}{r|}{0.2000} & 0.1957          \\ \hline
\multicolumn{1}{|l|}{Zapotec, Miahuatlán}               & zam-zamNT    & 0.1161 & \multicolumn{1}{r|}{0.1362} & 0.1369          \\ \hline
\multicolumn{1}{|l|}{Zapotec, Ozolotepec}               & zao-zaoNT    & 0.1810 & \multicolumn{1}{r|}{0.2056} & 0.2119          \\ \hline
\multicolumn{1}{|l|}{Zapotec, Rincón} &
  zar-zarNT* &
  0.1901 &
  \multicolumn{1}{l|}{N/A} &
  \multicolumn{1}{l|}{N/A} \\ \hline
\multicolumn{1}{|l|}{Zapotec, Santo Domingo Albarradas} & zas-zasNT    & 0.1858 & \multicolumn{1}{r|}{0.2301} & 02420           \\ \hline
\multicolumn{1}{|l|}{Zapotec, Tabaa} &
  zat-zatNTps* &
  \textbf{0.1869} &
  \multicolumn{1}{l|}{N/A} &
  \multicolumn{1}{l|}{N/A} \\ \hline
\multicolumn{1}{|l|}{Zapotec, Yatzachi}                 & zav-zavNT    & 0.1536 & \multicolumn{1}{r|}{0.1960} & 0.2049          \\ \hline
\multicolumn{1}{|l|}{Zapotec, Mitla}                    & zaw-zawNT    & 0.2094 & \multicolumn{1}{r|}{0.2202} & 0.2263          \\ \hline
\multicolumn{1}{|l|}{Zapotec, Coatecas Altas}           & zca-zcaNT    & 0.1567 & \multicolumn{1}{r|}{0.1855} & 0.1825          \\ \hline
\multicolumn{1}{|l|}{Zapotec, Choapan}                  & zpc-zpcNT    & 0.1439 & \multicolumn{1}{r|}{0.2057} & 0.2097          \\ \hline
\multicolumn{1}{|l|}{Zapotec, Mixtepec}                 & zpm-zpmNT    & 0.1215 & \multicolumn{1}{r|}{0.1387} & 0.1370          \\ \hline
\multicolumn{1}{|l|}{Zapotec, Amatlán}                  & zpo-zpoNT    & 0.1718 & \multicolumn{1}{r|}{0.1985} & 0.2027          \\ \hline
\multicolumn{1}{|l|}{Zapotec, Zoogocho}                 & zpq-zpqNT    & 0.1549 & \multicolumn{1}{r|}{0.2162} & 0.2371          \\ \hline
\multicolumn{1}{|l|}{Zapotec, Yalálag}                  & zpu-zpuNT    & 0.1557 & \multicolumn{1}{r|}{0.2148} & 0.2440          \\ \hline
\multicolumn{1}{|l|}{Zapotec, Chichicapan}              & zpv-zpvNT    & 0.1642 & \multicolumn{1}{r|}{0.1836} & 0.1948          \\ \hline
\multicolumn{1}{|l|}{Zapotec, Texmelucan}               & zpz-zpzNTpp  & 0.1438 & \multicolumn{1}{r|}{0.1727} & 0.1745          \\ \hline
\multicolumn{1}{|l|}{Zapotec, Southern Rincon}          & zsr-zsrNT    & 0.1988 & \multicolumn{1}{r|}{0.9000} & 0.3517          \\ \hline
\multicolumn{1}{|l|}{Zapotec, Quioquitani-Quierí}       & ztq-ztqNT    & 0.1676 & \multicolumn{1}{r|}{0.1944} & 0.2111          \\ \hline
\multicolumn{1}{|l|}{Zapotec, Yatee}                    & zty-ztyNTps* & 0.1782 & \multicolumn{1}{r|}{0.5623} & 0.4381          \\ \hline
\end{tabular}%
}
\caption{\label{tab:hmm_otomanguean}
Source / Target / Related Language Alignment Scores (Otomanguean).
}
\end{table}

\subsection{Sino-Tibetan}
For the Sino-Tibetan language family (458 total languages), the \emph{Tibeto-Burman > Western Tibeto-Burman} branch contains 442 languages.
Of these 442 languages, the eBible corpus contains 15 translations spread across the \emph{Kuki-Chin} (eight), \emph{Ngwi-Burmese} (three), and \emph{Western Tibeto-Burman} (four) sub-branches.
Translations from these sub-branches are for languages spoken in China, India, Myanmar, and Nepal, with eight full Bible, one \ac{NT}+, and fix \ac{NT}-only translations.
The best alignment results were seen with a Nepali translation (npi-npiulb) as the source and the Eastern Tamang translation (taj-taj) as the target.
The Limbu translation in Devanagari script (lif-lifNT2) was selected for the related language translation.
Results are summarized in Table \ref{tab:hmm_sino_tibetan}.

\begin{table}[]
\centering
\resizebox{0.49\textwidth}{!}{%
\begin{tabular}{|ll|rrrr|r|}
\hline
\multicolumn{2}{|c|}{\textbf{\begin{tabular}[c]{@{}c@{}}Target\\ Language(s)\end{tabular}}} &
  \multicolumn{4}{c|}{\textbf{\begin{tabular}[c]{@{}c@{}}National/Gateway\\ Language(s)\end{tabular}}} &
  \multicolumn{1}{c|}{\textbf{\begin{tabular}[c]{@{}c@{}}Related\\ Language(s)\end{tabular}}} \\ \hline
\multicolumn{1}{|c|}{\textbf{Language}} &
  \multicolumn{1}{c|}{\textbf{Translation}} &
  \multicolumn{1}{c|}{\textbf{hin-hin2017}} &
  \multicolumn{1}{c|}{\textbf{npi-npiulb}} &
  \multicolumn{1}{c|}{\textbf{mya-mya}} &
  \multicolumn{1}{c|}{\textbf{mya-myajvb}} &
  \multicolumn{1}{c|}{\textbf{taj-taj}} \\ \hline
\multicolumn{1}{|l|}{Zaiwa} &
  atb-atbNT &
  \multicolumn{1}{r|}{0.1827} &
  \multicolumn{1}{r|}{0.1640} &
  \multicolumn{1}{r|}{0.1621} &
  0.1703 &
  0.1545 \\ \hline
\multicolumn{1}{|l|}{Chin, Eastern Khumi} &
  cek-cekak &
  \multicolumn{1}{r|}{0.2112} &
  \multicolumn{1}{r|}{0.2074} &
  \multicolumn{1}{r|}{0.1259} &
  0.2237 &
  0.1655 \\ \hline
\multicolumn{1}{|l|}{Chin, Thaiphum} &
  cth-cth &
  \multicolumn{1}{r|}{0.1821} &
  \multicolumn{1}{r|}{0.1806} &
  \multicolumn{1}{r|}{0.1237} &
  0.2478 &
  0.1463 \\ \hline
\multicolumn{1}{|l|}{Chin, Siyin} &
  csy-csy &
  \multicolumn{1}{r|}{0.2078} &
  \multicolumn{1}{r|}{0.1843} &
  \multicolumn{1}{r|}{0.1423} &
  0.1854 &
  0.1590 \\ \hline
\multicolumn{1}{|l|}{Chin, Matu} &
  hlt-hlt &
  \multicolumn{1}{r|}{0.2281} &
  \multicolumn{1}{r|}{0.2170} &
  \multicolumn{1}{r|}{0.1229} &
  0.2208 &
  0.1620 \\ \hline
\multicolumn{1}{|l|}{Chin, Matu} &
  hlt-hltmcsb &
  \multicolumn{1}{r|}{0.2296} &
  \multicolumn{1}{r|}{0.2191} &
  \multicolumn{1}{r|}{0.1237} &
  0.2216 &
  0.1642 \\ \hline
\multicolumn{1}{|l|}{Chin, Matu} &
  hlt-hltthb &
  \multicolumn{1}{r|}{0.1678} &
  \multicolumn{1}{r|}{0.1698} &
  \multicolumn{1}{r|}{0.1195} &
  0.1559 &
  0.1449 \\ \hline
\multicolumn{1}{|l|}{Limbu (Limbu*)} &
  lif-lifNT &
  \multicolumn{1}{r|}{0.2108} &
  \multicolumn{1}{r|}{0.2665} &
  \multicolumn{1}{r|}{0.1142} &
  0.1513 &
  0.2664 \\ \hline
\multicolumn{1}{|l|}{Limbu (Deva)} &
  lif-lifNT2 &
  \multicolumn{1}{r|}{0.2093} &
  \multicolumn{1}{r|}{0.2660} &
  \multicolumn{1}{r|}{0.1131} &
  0.1509 &
  \textbf{0.2650} \\ \hline
\multicolumn{1}{|l|}{Sunwar} &
  suz-suzBl &
  \multicolumn{1}{r|}{0.2008} &
  \multicolumn{1}{r|}{0.2065} &
  \multicolumn{1}{r|}{0.1107} &
  0.1563 &
  0.1805 \\ \hline
\multicolumn{1}{|l|}{Tamang, Eastern} &
  taj-taj &
  \multicolumn{1}{r|}{0.1959} &
  \multicolumn{1}{r|}{\textbf{0.2657}} &
  \multicolumn{1}{r|}{0.1204} &
  0.1344 &
  \multicolumn{1}{l|}{N/A} \\ \hline
\multicolumn{1}{|l|}{Chin, Thado} &
  tcz-tczchongthu &
  \multicolumn{1}{r|}{0.1271} &
  \multicolumn{1}{r|}{0.1303} &
  \multicolumn{1}{r|}{0.0976} &
  0.1207 &
  0.1487 \\ \hline
\multicolumn{1}{|l|}{Chin, Zyphe} &
  zyp-zypNT &
  \multicolumn{1}{r|}{0.1765} &
  \multicolumn{1}{r|}{0.1824} &
  \multicolumn{1}{r|}{0.1399} &
  0.1583 &
  0.1662 \\ \hline
\end{tabular}%
}
\caption{\label{tab:hmm_sino_tibetan}
Source / Target / Related Language Alignment Scores (Sino-Tibetan).
}
\end{table}

\subsection{Trans-New Guinea}
For the Trans-New Guinea language family (481 total languages), the \emph{Finisterre-Huon > Finisterre} branch contains 40 languages and the \emph{Madang > Croisilles} branch contains 57 languages.
The eBible corpus contains 10 translations from the \emph{Finisterre-Huon > Finisterre} branch; five of these are \ac{NT}-only translations and five are \ac{NT}+ translations.  For the \emph{Madang > Croisilles} branch, the eBible corpus contains eight translations; six of these are \ac{NT}-only translations and two are \ac{NT}+ translations.
These translations are for languages spoken in Papua New Guinea, where national languages are English and Tok Pisin (tpi).
The best alignment results were observed with a Tok Pisin translation (tpi-tpiOTNT) as the source and the Yopno translation (yut-yut) as the target, while the Iyo translation (nca-nca) was selected for the related language translation.
Results are summarized in Table \ref{tab:hmm_trans_new_guinea}.

\begin{table}[]
\centering
\resizebox{0.49\textwidth}{!}{%
\begin{tabular}{|llrrll|}
\hline
\multicolumn{2}{|c|}{\textbf{\begin{tabular}[c]{@{}c@{}}Target\\ Language(s)\end{tabular}}} &
  \multicolumn{2}{c|}{\textbf{\begin{tabular}[c]{@{}c@{}}National/Gateway\\ Language(s)\end{tabular}}} &
  \multicolumn{2}{c|}{\textbf{\begin{tabular}[c]{@{}c@{}}Related\\ Language(s)\end{tabular}}} \\ \hline
\multicolumn{1}{|c|}{\textbf{Language}} &
  \multicolumn{1}{c|}{\textbf{Translation}} &
  \multicolumn{1}{c|}{\textbf{tpi-tpi}} &
  \multicolumn{1}{c|}{\textbf{tpi-tpiOTNT}} &
  \multicolumn{1}{c|}{\textbf{yut-yut}} &
  \multicolumn{1}{c|}{\textbf{aey-aey}} \\ \hline
\multicolumn{6}{|l|}{\textit{Finisterre-Huon > Finisterre Sub-branch}} \\ \hline
\multicolumn{1}{|l|}{Gwahatike} &
  \multicolumn{1}{l|}{dah-dah} & 
  \multicolumn{1}{r|}{0.1354} &
  \multicolumn{1}{r|}{0.1354} &
  \multicolumn{1}{r|}{0.1545} &
  N/A \\ \hline
\multicolumn{1}{|l|}{Tuma-Irumu} &
  \multicolumn{1}{l|}{iou-iou*} &
  \multicolumn{1}{r|}{0.1573} &
  \multicolumn{1}{r|}{0.1573} &
  \multicolumn{1}{r|}{0.1902} &
  N/A \\ \hline
\multicolumn{1}{|l|}{Iyo} &
  \multicolumn{1}{l|}{nca-nca} &
  \multicolumn{1}{r|}{0.2034} &
  \multicolumn{1}{r|}{0.2034} &
  \multicolumn{1}{r|}{\textbf{0.2677}} &
  N/A \\ \hline
\multicolumn{1}{|l|}{Numanggang} &
  \multicolumn{1}{l|}{nop-nop*} &
  \multicolumn{1}{r|}{0.1790} &
  \multicolumn{1}{r|}{0.1790} &
  \multicolumn{1}{r|}{0.2261} &
  N/A \\ \hline
\multicolumn{1}{|l|}{Rawa} &
  \multicolumn{1}{l|}{rwo-rwo-karo} &
  \multicolumn{1}{r|}{0.1913} &
  \multicolumn{1}{r|}{0.1913} &
  \multicolumn{1}{r|}{0.1859} &
  N/A \\ \hline
\multicolumn{1}{|l|}{Rawa} &
  \multicolumn{1}{l|}{rwo-rwo-rawa} &
  \multicolumn{1}{r|}{0.1917} &
  \multicolumn{1}{r|}{0.1917} &
  \multicolumn{1}{r|}{0.1854} &
  N/A \\ \hline
\multicolumn{1}{|l|}{Uri} &
  \multicolumn{1}{l|}{uvh-uvh} &
  \multicolumn{1}{r|}{0.1307} &
  \multicolumn{1}{r|}{0.1307} &
  \multicolumn{1}{r|}{0.1296} &
  N/A \\ \hline
\multicolumn{1}{|l|}{Wantoat} &
  \multicolumn{1}{l|}{wnc-wnc*} &
  \multicolumn{1}{r|}{0.1840} &
  \multicolumn{1}{r|}{0.1840} &
  \multicolumn{1}{r|}{0.1831} &
  N/A \\ \hline
\multicolumn{1}{|l|}{Yau} &
  \multicolumn{1}{l|}{yuw-yuw} &
  \multicolumn{1}{r|}{0.1774} &
  \multicolumn{1}{r|}{0.1774} &
  \multicolumn{1}{r|}{0.2161} &
  N/A \\ \hline
\multicolumn{1}{|l|}{Yopno} &
  \multicolumn{1}{l|}{yut-yut*} &
  \multicolumn{1}{r|}{\textbf{0.2252}} &
  \multicolumn{1}{r|}{\textbf{0.2252}} &
  \multicolumn{1}{l|}{N/A} &
  N/A \\ \hline
\multicolumn{6}{|l|}{\textit{Madang > Croisilles Sub-branch}} \\ \hline
\multicolumn{1}{|l|}{Amele} &
  \multicolumn{1}{l|}{aey-aey*} &
  \multicolumn{1}{r|}{\textbf{0.2089}} &
  \multicolumn{1}{r|}{\textbf{0.2089}} &
  \multicolumn{1}{l|}{N/A} &
  N/A \\ \hline
\multicolumn{1}{|l|}{Girawa} &
  \multicolumn{1}{l|}{bbr-bbr} &
  \multicolumn{1}{r|}{0.1948} &
  \multicolumn{1}{r|}{0.1948} &
  \multicolumn{1}{l|}{N/A} &
  \multicolumn{1}{r|}{0.2298} \\ \hline
\multicolumn{1}{|l|}{Nobonob} &
  \multicolumn{1}{l|}{gaw-gaw} &
  \multicolumn{1}{r|}{0.1875} &
  \multicolumn{1}{r|}{0.1875} &
  \multicolumn{1}{l|}{N/A} &
  \multicolumn{1}{r|}{\textbf{0.2432}} \\ \hline
\multicolumn{1}{|l|}{Kein} &
  \multicolumn{1}{l|}{bmh-bmh} &
  \multicolumn{1}{r|}{0.1934} &
  \multicolumn{1}{r|}{0.1934} &
  \multicolumn{1}{l|}{N/A} &
  \multicolumn{1}{r|}{0.2171} \\ \hline
\multicolumn{1}{|l|}{Mauwake} &
  \multicolumn{1}{l|}{mhl-mhl} &
  \multicolumn{1}{r|}{0.1659} &
  \multicolumn{1}{r|}{0.1659} &
  \multicolumn{1}{l|}{N/A} &
  \multicolumn{1}{r|}{0.1809} \\ \hline
\multicolumn{1}{|l|}{Bargam} &
  \multicolumn{1}{l|}{mlp-mlp} &
  \multicolumn{1}{r|}{0.1868} &
  \multicolumn{1}{r|}{0.1868} &
  \multicolumn{1}{l|}{N/A} &
  \multicolumn{1}{r|}{0.1971} \\ \hline
\multicolumn{1}{|l|}{Usan} &
  \multicolumn{1}{l|}{wnu-wnu} &
  \multicolumn{1}{r|}{0.1592} &
  \multicolumn{1}{r|}{0.1592} &
  \multicolumn{1}{l|}{N/A} &
  \multicolumn{1}{r|}{0.1765} \\ \hline
\multicolumn{1}{|l|}{Waskia} &
  \multicolumn{1}{l|}{wsk-wsk*} &
  \multicolumn{1}{r|}{0.1866} &
  \multicolumn{1}{r|}{0.1866} &
  \multicolumn{1}{l|}{N/A} &
  \multicolumn{1}{r|}{0.2353} \\ \hline
\end{tabular}%
}
\caption{\label{tab:hmm_trans_new_guinea}
Source / Target / Related Language Alignment Scores (Trans-New Guinea).
}
\end{table}

\section{Biblical Book Identifiers}
Table \ref{tab:usfm_codes} lists the standard three-character book identifiers utilized as part of the \ac{USFM} format.

\begin{table}[]
\centering
\resizebox{0.25\textwidth}{!}{%
\begin{tabular}{|c|c|}
\hline
\textbf{Book Name}   & \textbf{Identifier} \\ \hline
Genesis              & GEN                 \\ \hline
Exodus               & EXO                 \\ \hline
Leviticus            & LEV                 \\ \hline
Numbers              & NUM                 \\ \hline
Deuteronomy          & DEU                 \\ \hline
Joshua               & JOS                 \\ \hline
Judges               & JDG                 \\ \hline
Ruth                 & RUT                 \\ \hline
1 Samuel             & 1SA                 \\ \hline
2 Samuel             & 2SA                 \\ \hline
1 Kings              & 1KI                 \\ \hline
2 Kings              & 2KI                 \\ \hline
1 Chronicles         & 1CH                 \\ \hline
2 Chronicles         & 2CH                 \\ \hline
Ezra                 & EZR                 \\ \hline
Nehemiah             & NEH                 \\ \hline
Esther               & EST                 \\ \hline
Job                  & JOB                 \\ \hline
Psalms               & PSA                 \\ \hline
Proverbs             & PRO                 \\ \hline
Ecclesiastes         & ECC                 \\ \hline
Song of Songs        & SNG                 \\ \hline
Isaiah               & ISA                 \\ \hline
Jeremiah             & JER                 \\ \hline
Lamentations         & LAM                 \\ \hline
Ezekiel              & EZK                 \\ \hline
Daniel               & DAN                 \\ \hline
Hosea                & HOS                 \\ \hline
Joel                 & JOL                 \\ \hline
Amos                 & AMO                 \\ \hline
Obadiah              & OBA                 \\ \hline
Jonah                & JON                 \\ \hline
Micah                & MIC                 \\ \hline
Nahum                & NAH                 \\ \hline
Habakkuk             & HAB                 \\ \hline
Zephaniah            & ZEP                 \\ \hline
Haggai               & HAG                 \\ \hline
Zechariah            & ZEC                 \\ \hline
Malachi              & MAL                 \\ \hline
Matthew              & MAT                 \\ \hline
Mark                 & MRK                 \\ \hline
Luke                 & LUK                 \\ \hline
John                 & JHN                 \\ \hline
Acts of the Apostles & ACT                 \\ \hline
Romans               & ROM                 \\ \hline
1 Corinthians        & 1CO                 \\ \hline
2 Corinthians        & 2CO                 \\ \hline
Galatians            & GAL                 \\ \hline
Ephesians            & EPH                 \\ \hline
Philippians          & PHP                 \\ \hline
Colossians           & COL                 \\ \hline
1 Thessalonians      & 1TH                 \\ \hline
2 Thessalonians      & 2TH                 \\ \hline
1 Timothy            & 1TI                 \\ \hline
2 Timothy            & 2TI                 \\ \hline
Titus                & TIT                 \\ \hline
Philemon             & PHM                 \\ \hline
Hebrews              & HEB                 \\ \hline
James                & JAS                 \\ \hline
1 Peter              & 1PE                 \\ \hline
2 Peter              & 2PE                 \\ \hline
1 John               & 1JN                 \\ \hline
2 John               & 2JN                 \\ \hline
3 John               & 3JN                 \\ \hline
Jude                 & JUD                 \\ \hline
Revelation           & REV                 \\ \hline
\end{tabular}%
}
\caption{\label{tab:usfm_codes}
Three-letter identifiers for Biblical books (excluding those not used in this work).
}
\end{table}

\section{Extended Results}
In this section, we include additional results of interest, represented in tabular form. Several of these tables have figure analogs in the main body of the paper.
First, we consider the effect of increasing model size for Meta’s \ac{NLLB} model, which is available in five different configurations, ranging from 600M to 54.5B parameters.
To evaluate the potential benefits of working with the larger models, the \ac{NLLB}-1.3B (distilled) model was fine-tuned on one of the same train/test/validation splits from the previous tests.
This was done for the Dravidian, Niger-Congo, and Sino-Tibetan translation pairings.
Translation accuracy metrics with the \ac{NLLB}-1.3B-distilled models were consistently better for each translation pairing than with the \ac{NLLB}-600M model, as shown in \ref{tab:nllb_size}.

\begin{table}[]
\centering
\resizebox{0.49\textwidth}{!}{%
\begin{tabular}{|c|c|c|c|c|c|c|}
\hline
\textbf{TranslationPairing} &
  \textbf{Model} &
  \textbf{BLEU} &
  \textbf{spBLEU} &
  \textbf{chrF3} &
  \textbf{WER} &
  \textbf{TER} \\ \hline
\multirow{2}{*}{Dravidian} &
  NLLB-600M &
  21.7 &
  40.0 &
  59.5 &
  46.6 &
  69.7 \\ \cline{2-7} 
 &
  NLLB-1.3B-distilled &
  \textbf{\begin{tabular}[c]{@{}c@{}}24.9 \\ (+3.2)\end{tabular}} &
  \textbf{\begin{tabular}[c]{@{}c@{}}44.2 \\ (+4.2)\end{tabular}} &
  \textbf{\begin{tabular}[c]{@{}c@{}}62.7 \\ (+3.2)\end{tabular}} &
  \textbf{\begin{tabular}[c]{@{}c@{}}43.9 \\ (-2.7)\end{tabular}} &
  \textbf{\begin{tabular}[c]{@{}c@{}}65.6 \\ (-4.1)\end{tabular}} \\ \hline
\multirow{2}{*}{Niger-Congo} &
  NLLB-600M &
  28.3 &
  37.9 &
  60.5 &
  43.6 &
  65.2 \\ \cline{2-7} 
 &
  NLLB-1.3B-distilled &
  \textbf{\begin{tabular}[c]{@{}c@{}}29.9 \\ (+1.6)\end{tabular}} &
  \textbf{\begin{tabular}[c]{@{}c@{}}40.0 \\ (+2.1)\end{tabular}} &
  \textbf{\begin{tabular}[c]{@{}c@{}}62.2 \\ (+1.7)\end{tabular}} &
  \textbf{\begin{tabular}[c]{@{}c@{}}41.7 \\ (-1.9)\end{tabular}} &
  \textbf{\begin{tabular}[c]{@{}c@{}}63.1 \\ (-2.1)\end{tabular}} \\ \hline
\multirow{2}{*}{Sino-Tibetan} &
  NLLB-600M &
  31.5 &
  49.5 &
  57.9 &
  \textbf{52.0} &
  64.0 \\ \cline{2-7} 
 &
  NLLB-1.3B-distilled &
  \textbf{\begin{tabular}[c]{@{}c@{}}33.1 \\ (+1.6)\end{tabular}} &
  \textbf{\begin{tabular}[c]{@{}c@{}}51.0 \\ (+1.5)\end{tabular}} &
  \textbf{\begin{tabular}[c]{@{}c@{}}59.9 \\ (+2.0)\end{tabular}} &
  \begin{tabular}[c]{@{}c@{}}52.1 \\ (+0.1)\end{tabular} &
  \textbf{\begin{tabular}[c]{@{}c@{}}62.3 \\ (-1.7)\end{tabular}} \\ \hline
\end{tabular}%
}
\caption{\label{tab:nllb_size}
Model size impact, \ac{NLLB}-600M versus \ac{NLLB}-1.3B (distilled).
}
\end{table}

Next, we consider how model complexity improves performance across all five scoring metrics, as shown in Table \ref{tab:trans_across_types}.

\begin{table}[]
\centering
\resizebox{0.49\textwidth}{!}{%
\begin{tabular}{|c|c|c|c|c|c|c|}
\hline
\textbf{TranslationPairing} & \textbf{MT Technology} & \textbf{BLEU} & \textbf{spBLEU} & \textbf{chrF3} & \textbf{WER} & \textbf{TER} \\ \hline
\multirow{4}{*}{Dravidian}    & SMT                  & 9.8           & 20.1          & 36.9          & 98.6          & 84.4          \\ \cline{2-7} 
                              & OpenNMT              & 13.3          & 28.0          & 46.9          & 56.3          & 79.0          \\ \cline{2-7} 
                              & NLLB-600M            & 21.7          & 39.8          & 58.2          & 47.0          & 68.6          \\ \cline{2-7} 
                              & NLLB-1.3B-distilled* & \textbf{24.9} & \textbf{44.2} & \textbf{62.7} & \textbf{43.9} & \textbf{65.6} \\ \hline
\multirow{4}{*}{Niger-Congo}  & SMT                  & 19.3          & 27.6          & 49.2          & 81.6          & 73.8          \\ \cline{2-7} 
                              & OpenNMT              & 16.6          & 24.3          & 45.2          & 55.2          & 77.9          \\ \cline{2-7} 
                              & NLLB-600M            & 28.8          & 37.9          & 60.4          & 43.9          & 65.2          \\ \cline{2-7} 
                              & NLLB-1.3B-distilled* & \textbf{29.9} & \textbf{40.0} & \textbf{62.2} & \textbf{41.7} & \textbf{63.1} \\ \hline
\multirow{4}{*}{Sino-Tibetan} & SMT                  & 9.9           & 30.8          & 42.1          & 60.2          & 84.4          \\ \cline{2-7} 
                              & OpenNMT              & 10.2          & 26.5          & 37.4          & 66.9          & 86.8          \\ \cline{2-7} 
                              & NLLB-600M            & 31.5          & 49.5          & 57.9          & \textbf{52.0} & 64.0          \\ \cline{2-7} 
                              & NLLB-1.3B-distilled* & \textbf{33.1} & \textbf{51.0} & \textbf{59.9} & 52.1          & \textbf{62.3} \\ \hline
\end{tabular}%
}
\caption{\label{tab:trans_across_types}
Machine translation across model types (\ac{SMT}, OpenNMT, and \ac{NLLB}-600M).
*Five-fold \ac{CV} is not performed on \ac{NLLB}-1.3B-distilled due to training overheads.
}
\end{table}

We also reiterate the median performance across scoring metrics, along with counts for the total number of available verses per each target translation, as shown in Table \ref{tab:pairings_acc}.

\begin{table}[]
\centering
\resizebox{0.49\textwidth}{!}{%
\begin{tabular}{|c|c|c|c|c|c|c|}
\hline
\textbf{TranslationPairing} & \textbf{Training Verses} & \textbf{BLEU} & \textbf{spBLEU} & \textbf{chrF3} & \textbf{WER}      & \textbf{TER}      \\ \hline
Afro-Asiatic     & 7,394  & 29.9±1.4          & 38.4±1.5          & 52.5±1.2          & 51.6±1.3 & 67.2±1.6 \\ \hline
Austronesian     & 7,404  & \textbf{35.1±1.1} & 39.4±1.1          & 54.2±0.6          & 49.1±1.6 & 62.0±1.4 \\ \hline
Dravidian        & 30,589 & 21.7±0.8          & 39.8±0.8          & 58.2±0.6          & 47.0±0.9 & 68.6±0.9 \\ \hline
Indo-European               & 30,599                   & 30.5±1.0      & 40.4±1.1        & 55.7±0.7       & \textbf{42.5±1.1} & \textbf{55.8±1.4} \\ \hline
Niger-Congo      & 13,304 & 28.8±1.0          & 37.9±0.7          & \textbf{60.4±0.6} & 43.9±0.7 & 65.2±1.1 \\ \hline
Otomanguean      & 9,901  & 28.3±0.9          & 44.3±0.8          & 56.4±1.2          & 50.9±0.6 & 68.1±0.7 \\ \hline
Sino-Tibetan     & 7,419  & 31.5±1.6          & \textbf{49.5±1.5} & 58.9±0.7          & 50.5±1.3 & 63.9±1.7 \\ \hline
Trans-New Guinea & 9,775  & 31.6±0.7          & 49.0±0.5          & 60.1±0.9          & 48.0±0.8 & 61.6±0.4 \\ \hline
\end{tabular}%
}
\caption{\label{tab:pairings_acc}
Median translation accuracy scores by translation pairing.
}
\end{table}

Table \ref{tab:nllb_fine_tune_GT} shows BLEU, spBLEU, and chrF3 scores for the \emph{CV} task as well as the \emph{Gospel Translation} task.

\begin{table}[]
\centering
\resizebox{0.49\textwidth}{!}{%
\begin{tabular}{|c|ccc|ccc|}
\hline
\multicolumn{1}{|l|}{} &
  \multicolumn{3}{c|}{\textbf{CV Task}} &
  \multicolumn{3}{c|}{\textbf{Gospel Translation Task}} \\ \hline
\textbf{TranslationPairing} &
  \multicolumn{1}{c|}{\textbf{BLEU}} &
  \multicolumn{1}{c|}{\textbf{spBLEU}} &
  \textbf{chrF3} &
  \multicolumn{1}{c|}{\textbf{BLEU}} &
  \multicolumn{1}{c|}{\textbf{spBLEU}} &
  \textbf{chrF3} \\ \hline
Afro-Asiatic &
  \multicolumn{1}{c|}{29.9} &
  \multicolumn{1}{c|}{\textbf{38.4}} &
  \textbf{52.1} &
  \multicolumn{1}{c|}{\textbf{30.2 (+0.3)}} &
  \multicolumn{1}{c|}{35.8 (-2.6)} &
  49.7 (-2.4) \\ \hline
Austronesian &
  \multicolumn{1}{c|}{\textbf{35.1}} &
  \multicolumn{1}{c|}{\textbf{39.4}} &
  \textbf{53.8} &
  \multicolumn{1}{c|}{33.7 (-1.4)} &
  \multicolumn{1}{c|}{36.2 (-3.2)} &
  52.6 (-1.2) \\ \hline
Dravidian &
  \multicolumn{1}{c|}{\textbf{21.7}} &
  \multicolumn{1}{c|}{\textbf{39.8}} &
  \textbf{58.5} &
  \multicolumn{1}{c|}{18.2 (-3.5)} &
  \multicolumn{1}{c|}{35.1 (-4.7)} &
  58.1 (-0.4) \\ \hline
Indo-European &
  \multicolumn{1}{c|}{30.5} &
  \multicolumn{1}{c|}{40.4} &
  55.5 &
  \multicolumn{1}{c|}{\textbf{36.5 (+6.0)}} &
  \multicolumn{1}{c|}{\textbf{44.5 (+4.1)}} &
  \textbf{59.9 (+4.4)} \\ \hline
Niger-Congo &
  \multicolumn{1}{c|}{\textbf{28.8}} &
  \multicolumn{1}{c|}{\textbf{37.9}} &
  \textbf{60.1} &
  \multicolumn{1}{c|}{21.7 (-7.1)} &
  \multicolumn{1}{c|}{29.7 (-10.2)} &
  53.7 (-6.4) \\ \hline
Otomanguean &
  \multicolumn{1}{c|}{\textbf{28.3}} &
  \multicolumn{1}{c|}{\textbf{44.3}} &
  \textbf{56.7} &
  \multicolumn{1}{c|}{20.4 (-7.9)} &
  \multicolumn{1}{c|}{34.5 (-9.8)} &
  48.7 (-8.0) \\ \hline
Sino-Tibetan &
  \multicolumn{1}{c|}{\textbf{31.5}} &
  \multicolumn{1}{c|}{\textbf{49.5}} &
  \textbf{58.8} &
  \multicolumn{1}{c|}{30.6 (-0.9)} &
  \multicolumn{1}{c|}{45.9 (-3.6)} &
  56.1 (-2.7) \\ \hline
Trans-New Guinea &
  \multicolumn{1}{c|}{\textbf{31.6}} &
  \multicolumn{1}{c|}{\textbf{49.0}} &
  \textbf{59.8} &
  \multicolumn{1}{c|}{28.5 (-3.1)} &
  \multicolumn{1}{c|}{41.7 (-7.3)} &
  51.7 (-8.1) \\ \hline
\end{tabular}%
}
\caption{\label{tab:nllb_fine_tune_GT}
Scores for \ac{NLLB}-600M fine-tuned models for the \emph{\ac{CV}} task and the \emph{Gospel Translation} task.
}
\end{table}

Table \ref{tab:nllb_fine_tune_EOT} shows BLEU scores for various minor-prophet books with both the \emph{Early \ac{OT}} task (training set includes the full \ac{NT}) and the \emph{Late \ac{OT}} task (training set includes the full \ac{NT} and \ac{OT} books except the minor prophets in the test set, shown below).

\begin{table}[]
\centering
\resizebox{0.49\textwidth}{!}{%
\begin{tabular}{|c|cc|cc|}
\hline
                       & \multicolumn{2}{c|}{\textbf{Dravidian (BLEU)}}                      & \multicolumn{2}{c|}{\textbf{Indo-European (BLEU)}}                  \\ \hline
\textbf{Minor Prophet} & \multicolumn{1}{c|}{\textbf{Early OT Task}} & \textbf{Late OT Task} & \multicolumn{1}{c|}{\textbf{Early OT Task}} & \textbf{Late OT Task} \\ \hline
HOS & \multicolumn{1}{c|}{10.4} & 16.6 (+6.2)  & \multicolumn{1}{c|}{16.6} & 22.6 (+6.0)  \\ \hline
JOL & \multicolumn{1}{c|}{10.2} & 19.2 (+9.0)  & \multicolumn{1}{c|}{19.2} & 28.5 (+9.3)  \\ \hline
AMO & \multicolumn{1}{c|}{7.2}  & 17.9 (+10.7) & \multicolumn{1}{c|}{19.0} & 26.0 (+7.0)  \\ \hline
OBA & \multicolumn{1}{c|}{5.4}  & 16.7 (+11.3) & \multicolumn{1}{c|}{16.9} & 26.3 (+9.4)  \\ \hline
MIC & \multicolumn{1}{c|}{7.0}  & 14.7 (+7.7)  & \multicolumn{1}{c|}{20.8} & 25.8 (+5.0)  \\ \hline
NAH & \multicolumn{1}{c|}{4.5}  & 10.9 (+6.4)  & \multicolumn{1}{c|}{13.8} & 21.4 (+7.6)  \\ \hline
HAB & \multicolumn{1}{c|}{5.9}  & 9.8 (+3.9)   & \multicolumn{1}{c|}{18.9} & 22.2 (+3.3)  \\ \hline
ZEP & \multicolumn{1}{c|}{8.1}  & 15.1 (+7.0)  & \multicolumn{1}{c|}{19.1} & 29.2 (+10.1) \\ \hline
HAG & \multicolumn{1}{c|}{6.9}  & 21.3 (+14.4) & \multicolumn{1}{c|}{17.9} & 34.3 (+16.4) \\ \hline
ZEC & \multicolumn{1}{c|}{8.3}  & 16.6 (+8.3)  & \multicolumn{1}{c|}{17.1} & 27.1 (+10.0) \\ \hline
MAL & \multicolumn{1}{c|}{8.2}  & 11.9 (+3.7)  & \multicolumn{1}{c|}{15.4} & 22.8 (+7.4)  \\ \hline
\end{tabular}%
}
\caption{\label{tab:nllb_fine_tune_EOT}
BLEU scores for \ac{NLLB}-600M fine-tuned models on \emph{Early \ac{OT}} and \emph{Late \ac{OT}} tasks.
}
\end{table}

Table \ref{tab:nllb_fine_tune_EOT_book} shows BLEU scores for the \ac{NLLB} model on the \emph{Early \ac{OT}} task.

\begin{table}[]
\centering
\resizebox{0.49\textwidth}{!}{%
\begin{tabular}{|c|c|c|c|c|c|c|c|c|}
\hline
\textbf{TranslationPairing} &
  \textbf{GEN} &
  \textbf{EXO} &
  \textbf{LEV} &
  \textbf{NUM} &
  \textbf{DEU} &
  \textbf{RUT} &
  \textbf{PSA} &
  \textbf{JON} \\ \hline
Dravidian &
  11.8 &
  8.9 &
  6.4 &
  6.5 &
  6.4 &
  8.1 &
  10.9 &
  9.2 \\ \hline
Indo-European &
  23.6 &
  17.0 &
  19.3 &
  16.7 &
  17.9 &
  18.6 &
  20.3 &
  20.9 \\ \hline
Niger-Congo &
  18.9 &
  16.0 &
  12.7 &
  14.7 &
  12.8 &
  N/A &
  N/A &
  N/A \\ \hline
Otomanguean &
  N/A &
  N/A &
  N/A &
  N/A &
  N/A &
  N/A &
  16.2 &
  N/A \\ \hline
Trans-New Guinea &
  \multicolumn{1}{l|}{N/A} &
  \multicolumn{1}{l|}{N/A} &
  \multicolumn{1}{l|}{N/A} &
  \multicolumn{1}{l|}{N/A} &
  \multicolumn{1}{l|}{N/A} &
  \multicolumn{1}{l|}{N/A} &
  19.0 &
  \multicolumn{1}{l|}{N/A} \\ \hline
\end{tabular}%
}
\caption{\label{tab:nllb_fine_tune_EOT_book}
BLEU scores for \ac{NLLB}-600M fine-tuned models on the \emph{Early \ac{OT}} task.
}
\end{table}

Table \ref{tab:nllb_related_lang} shows various scoring metrics for the \emph{Gospel Translation} task, both with and without related languages supplied.

\begin{table}[]
\centering
\resizebox{0.49\textwidth}{!}{%
\begin{tabular}{|c|ccc|ccc|}
\hline
\multirow{2}{*}{\begin{tabular}[c]{@{}c@{}}Translation\\ Pairing\end{tabular}} &
  \multicolumn{3}{c|}{\textbf{\begin{tabular}[c]{@{}c@{}}Gospel Translation (MAT)\\ (Without Related Language)\end{tabular}}} &
  \multicolumn{3}{c|}{\textbf{\begin{tabular}[c]{@{}c@{}}Gospel Translation (MAT)\\ (With Related Language)\end{tabular}}} \\ \cline{2-7} 
 &
  \multicolumn{1}{c|}{\textbf{BLEU}} &
  \multicolumn{1}{c|}{\textbf{spBLEU}} &
  \textbf{chrF3} &
  \multicolumn{1}{c|}{\textbf{BLEU}} &
  \multicolumn{1}{c|}{\textbf{spBLEU}} &
  \textbf{chrF3} \\ \hline
Afro-Asiatic &
  \multicolumn{1}{c|}{\textbf{30.2}} &
  \multicolumn{1}{c|}{\textbf{35.8}} &
  \textbf{49.7} &
  \multicolumn{1}{c|}{28.5} &
  \multicolumn{1}{c|}{33.8} &
  48.0 \\ \hline
Austronesian &
  \multicolumn{1}{c|}{33.7} &
  \multicolumn{1}{c|}{36.2} &
  52.6 &
  \multicolumn{1}{c|}{\textbf{36.2}} &
  \multicolumn{1}{c|}{\textbf{39.0}} &
  \textbf{54.2} \\ \hline
Dravidian &
  \multicolumn{1}{c|}{18.2} &
  \multicolumn{1}{c|}{35.1} &
  58.1 &
  \multicolumn{1}{c|}{\textbf{19.4}} &
  \multicolumn{1}{c|}{\textbf{36.5}} &
  \textbf{58.7} \\ \hline
Indo-European &
  \multicolumn{1}{c|}{\textbf{36.5}} &
  \multicolumn{1}{c|}{\textbf{44.5}} &
  59.9 &
  \multicolumn{1}{c|}{36.0} &
  \multicolumn{1}{c|}{44.2} &
  \textbf{60.1} \\ \hline
Niger-Congo &
  \multicolumn{1}{c|}{\textbf{21.7}} &
  \multicolumn{1}{c|}{\textbf{29.7}} &
  \textbf{53.7} &
  \multicolumn{1}{c|}{21.1} &
  \multicolumn{1}{c|}{28.7} &
  52.2 \\ \hline
Otomanguean &
  \multicolumn{1}{c|}{\textbf{20.4}} &
  \multicolumn{1}{c|}{\textbf{34.5}} &
  \textbf{48.7} &
  \multicolumn{1}{c|}{18.1} &
  \multicolumn{1}{c|}{31.2} &
  47.9 \\ \hline
Sino-Tibetan &
  \multicolumn{1}{c|}{\textbf{30.6}} &
  \multicolumn{1}{c|}{\textbf{45.9}} &
  \textbf{56.1} &
  \multicolumn{1}{c|}{27.3} &
  \multicolumn{1}{c|}{41.6} &
  53.9 \\ \hline
Trans-New Guinea &
  \multicolumn{1}{c|}{\textbf{28.5}} &
  \multicolumn{1}{c|}{\textbf{41.7}} &
  \textbf{51.7} &
  \multicolumn{1}{c|}{27.8} &
  \multicolumn{1}{c|}{40.5} &
  51.3 \\ \hline
\end{tabular}%
}
\caption{\label{tab:nllb_related_lang}
BLEU, spBLEU, and chrF3 scores for \ac{NLLB}-600M fine-tuned models on \emph{Gospel Translation} (without \emph{Related Language}) task and the \emph{Gospel Translation} (with \emph{Related Language}) task.
}
\end{table}

Table \ref{tab:nllb_epistles} shows the performance of fine-tuned \ac{NLLB} on the \emph{Epistle Translation} task.

\begin{table}[]
\centering
\resizebox{0.49\textwidth}{!}{%
\begin{tabular}{|c|ccccc|ccccc|}
\hline
\multirow{2}{*}{\textbf{\begin{tabular}[c]{@{}c@{}}Translation\\ Pairing\end{tabular}}} &
  \multicolumn{5}{c|}{\textbf{\begin{tabular}[c]{@{}c@{}}Epistle Translation\\ (Without Related Language)\end{tabular}}} &
  \multicolumn{5}{c|}{\textbf{\begin{tabular}[c]{@{}c@{}}Epistle Translation\\ (With Related Language)\end{tabular}}} \\ \cline{2-11} 
 &
  \multicolumn{1}{c|}{\textbf{1TH}} &
  \multicolumn{1}{c|}{\textbf{2TH}} &
  \multicolumn{1}{c|}{\textbf{1TI}} &
  \multicolumn{1}{c|}{\textbf{2TI}} &
  \textbf{TIT} &
  \multicolumn{1}{c|}{\textbf{1TH}} &
  \multicolumn{1}{c|}{\textbf{2TH}} &
  \multicolumn{1}{c|}{\textbf{1TI}} &
  \multicolumn{1}{c|}{\textbf{2TI}} &
  \textbf{TIT} \\ \hline
Afro-Asiatic &
  \multicolumn{1}{c|}{\textbf{14.8}} &
  \multicolumn{1}{c|}{\textbf{17.3}} &
  \multicolumn{1}{c|}{\textbf{11.3}} &
  \multicolumn{1}{c|}{\textbf{14.3}} &
  \textbf{9.3} &
  \multicolumn{1}{c|}{14.2} &
  \multicolumn{1}{c|}{14.6} &
  \multicolumn{1}{c|}{11.2} &
  \multicolumn{1}{c|}{12.3} &
  8.1 \\ \hline
Austronesian &
  \multicolumn{1}{c|}{20.1} &
  \multicolumn{1}{c|}{21.5} &
  \multicolumn{1}{c|}{19.9} &
  \multicolumn{1}{c|}{23.5} &
  19.8 &
  \multicolumn{1}{c|}{\textbf{30.0}} &
  \multicolumn{1}{c|}{\textbf{31.3}} &
  \multicolumn{1}{c|}{\textbf{30.5}} &
  \multicolumn{1}{c|}{\textbf{33.4}} &
  \textbf{28.3} \\ \hline
Dravidian &
  \multicolumn{1}{c|}{11.4} &
  \multicolumn{1}{c|}{\textbf{13.4}} &
  \multicolumn{1}{c|}{7.0} &
  \multicolumn{1}{c|}{8.1} &
  \textbf{5.2} &
  \multicolumn{1}{c|}{\textbf{13.2}} &
  \multicolumn{1}{c|}{13.1} &
  \multicolumn{1}{c|}{\textbf{7.5}} &
  \multicolumn{1}{c|}{\textbf{9.4}} &
  4.8 \\ \hline
Indo-European &
  \multicolumn{1}{c|}{\textbf{28.7}} &
  \multicolumn{1}{c|}{\textbf{26.2}} &
  \multicolumn{1}{c|}{\textbf{21.3}} &
  \multicolumn{1}{c|}{\textbf{26.7}} &
  \textbf{21.6} &
  \multicolumn{1}{c|}{26.0} &
  \multicolumn{1}{c|}{24.0} &
  \multicolumn{1}{c|}{18.8} &
  \multicolumn{1}{c|}{24.2} &
  20.2 \\ \hline
Niger-Congo &
  \multicolumn{1}{c|}{14.9} &
  \multicolumn{1}{c|}{16.7} &
  \multicolumn{1}{c|}{15.1} &
  \multicolumn{1}{c|}{14.5} &
  13.6 &
  \multicolumn{1}{c|}{\textbf{15.5}} &
  \multicolumn{1}{c|}{\textbf{18.4}} &
  \multicolumn{1}{c|}{\textbf{17.1}} &
  \multicolumn{1}{c|}{\textbf{18.6}} &
  \textbf{16.0} \\ \hline
Otomanguean &
  \multicolumn{1}{c|}{\textbf{16.2}} &
  \multicolumn{1}{c|}{\textbf{14.0}} &
  \multicolumn{1}{c|}{\textbf{15.5}} &
  \multicolumn{1}{c|}{13.1} &
  \textbf{15.6} &
  \multicolumn{1}{c|}{14.2} &
  \multicolumn{1}{c|}{13.1} &
  \multicolumn{1}{c|}{15.4} &
  \multicolumn{1}{c|}{\textbf{13.4}} &
  14.0 \\ \hline
Sino-Tibetan &
  \multicolumn{1}{c|}{\textbf{18.8}} &
  \multicolumn{1}{c|}{17.2} &
  \multicolumn{1}{c|}{\textbf{16.2}} &
  \multicolumn{1}{c|}{\textbf{18.2}} &
  \textbf{18.3} &
  \multicolumn{1}{c|}{18.3} &
  \multicolumn{1}{c|}{\textbf{18.9}} &
  \multicolumn{1}{c|}{15.7} &
  \multicolumn{1}{c|}{16.6} &
  14.4 \\ \hline
Trans-New Guinea &
  \multicolumn{1}{c|}{\textbf{18.7}} &
  \multicolumn{1}{c|}{\textbf{20.5}} &
  \multicolumn{1}{c|}{\textbf{16.9}} &
  \multicolumn{1}{c|}{\textbf{17.6}} &
  \textbf{15.0} &
  \multicolumn{1}{c|}{16.8} &
  \multicolumn{1}{c|}{20.4} &
  \multicolumn{1}{c|}{13.7} &
  \multicolumn{1}{c|}{13.6} &
  14.8 \\ \hline
\end{tabular}%
}
\caption{\label{tab:nllb_epistles}
BLEU scores for \ac{NLLB}-600M fine-tuned models on \emph{Epistle Translation} (without \emph{Related Language}) task and the \emph{Epistle Translation} (with \emph{Related Language}) task.
}
\end{table}

Table \ref{tab:nllb_NT_completion} shows the fine-tuned \ac{NLLB} model performance on the \emph{\ac{NT} Completion} task.

\begin{table}[]
\centering
\resizebox{0.49\textwidth}{!}{%
\begin{tabular}{|c|cccccc|cccccc|}
\hline
\multirow{3}{*}{\begin{tabular}[c]{@{}c@{}}Translation\\ Pairing\end{tabular}} &
  \multicolumn{6}{c|}{\textbf{\begin{tabular}[c]{@{}c@{}}NT Completion\\ (Without Related Language)\end{tabular}}} &
  \multicolumn{6}{c|}{\textbf{\begin{tabular}[c]{@{}c@{}}NT Completion\\ (With Related Language)\end{tabular}}} \\ \cline{2-13} 
 &
  \multicolumn{3}{c|}{\textbf{ROM}} &
  \multicolumn{3}{c|}{\textbf{REV}} &
  \multicolumn{3}{c|}{\textbf{ROM}} &
  \multicolumn{3}{c|}{\textbf{REV}} \\ \cline{2-13} 
 &
  \multicolumn{1}{c|}{\textbf{BLEU}} &
  \multicolumn{1}{c|}{\textbf{spBLEU}} &
  \multicolumn{1}{c|}{\textbf{chrF3}} &
  \multicolumn{1}{c|}{\textbf{BLEU}} &
  \multicolumn{1}{c|}{\textbf{spBLEU}} &
  \textbf{chrF3} &
  \multicolumn{1}{c|}{\textbf{BLEU}} &
  \multicolumn{1}{c|}{\textbf{spBLEU}} &
  \multicolumn{1}{c|}{\textbf{chrF3}} &
  \multicolumn{1}{c|}{\textbf{BLEU}} &
  \multicolumn{1}{c|}{\textbf{spBLEU}} &
  \textbf{chrF3} \\ \hline
Afro-Asiatic &
  \multicolumn{1}{c|}{18.9} &
  \multicolumn{1}{c|}{28.2} &
  \multicolumn{1}{c|}{43.7} &
  \multicolumn{1}{c|}{16.0} &
  \multicolumn{1}{c|}{\textbf{25.6}} &
  43.3 &
  \multicolumn{1}{c|}{\textbf{19.4}} &
  \multicolumn{1}{c|}{\textbf{29.1}} &
  \multicolumn{1}{c|}{\textbf{44.3}} &
  \multicolumn{1}{c|}{\textbf{16.2}} &
  \multicolumn{1}{c|}{25.1} &
  \textbf{43.7} \\ \hline
Austronesian &
  \multicolumn{1}{c|}{23.2} &
  \multicolumn{1}{c|}{27.3} &
  \multicolumn{1}{c|}{44.9} &
  \multicolumn{1}{c|}{28.6} &
  \multicolumn{1}{c|}{33.3} &
  50.3 &
  \multicolumn{1}{c|}{\textbf{31.5}} &
  \multicolumn{1}{c|}{\textbf{35.0}} &
  \multicolumn{1}{c|}{\textbf{50.7}} &
  \multicolumn{1}{c|}{\textbf{40.1}} &
  \multicolumn{1}{c|}{\textbf{44.0}} &
  \textbf{58.0} \\ \hline
Dravidian &
  \multicolumn{1}{c|}{\textbf{12.8}} &
  \multicolumn{1}{c|}{29.3} &
  \multicolumn{1}{c|}{53.1} &
  \multicolumn{1}{c|}{13.7} &
  \multicolumn{1}{c|}{31.9} &
  54.6 &
  \multicolumn{1}{c|}{12.6} &
  \multicolumn{1}{c|}{\textbf{29.5}} &
  \multicolumn{1}{c|}{53.1} &
  \multicolumn{1}{c|}{\textbf{14.6}} &
  \multicolumn{1}{c|}{\textbf{32.6}} &
  \textbf{55.7} \\ \hline
Indo-European &
  \multicolumn{1}{c|}{\textbf{30.0}} &
  \multicolumn{1}{c|}{\textbf{38.2}} &
  \multicolumn{1}{c|}{\textbf{56.1}} &
  \multicolumn{1}{c|}{\textbf{29.2}} &
  \multicolumn{1}{c|}{\textbf{38.1}} &
  \textbf{54.3} &
  \multicolumn{1}{c|}{29.1} &
  \multicolumn{1}{c|}{37.9} &
  \multicolumn{1}{c|}{55.5} &
  \multicolumn{1}{c|}{28.4} &
  \multicolumn{1}{c|}{37.7} &
  54.2 \\ \hline
Niger-Congo &
  \multicolumn{1}{c|}{20.1} &
  \multicolumn{1}{c|}{29.9} &
  \multicolumn{1}{c|}{54.6} &
  \multicolumn{1}{c|}{\textbf{23.5}} &
  \multicolumn{1}{c|}{\textbf{32.3}} &
  55.3 &
  \multicolumn{1}{c|}{\textbf{22.3}} &
  \multicolumn{1}{c|}{\textbf{32.1}} &
  \multicolumn{1}{c|}{\textbf{57.1}} &
  \multicolumn{1}{c|}{23.4} &
  \multicolumn{1}{c|}{31.1} &
  \textbf{56.2} \\ \hline
Otomanguean &
  \multicolumn{1}{c|}{22.1} &
  \multicolumn{1}{c|}{\textbf{38.8}} &
  \multicolumn{1}{c|}{53.1} &
  \multicolumn{1}{c|}{20.6} &
  \multicolumn{1}{c|}{35.5} &
  51.1 &
  \multicolumn{1}{c|}{22.1} &
  \multicolumn{1}{c|}{38.6} &
  \multicolumn{1}{c|}{\textbf{53.5}} &
  \multicolumn{1}{c|}{\textbf{21.2}} &
  \multicolumn{1}{c|}{\textbf{36.0}} &
  \textbf{52.3} \\ \hline
Sino-Tibetan &
  \multicolumn{1}{c|}{23.3} &
  \multicolumn{1}{c|}{41.4} &
  \multicolumn{1}{c|}{51.6} &
  \multicolumn{1}{c|}{23.3} &
  \multicolumn{1}{c|}{\textbf{42.5}} &
  53.2 &
  \multicolumn{1}{c|}{\textbf{23.9}} &
  \multicolumn{1}{c|}{\textbf{41.5}} &
  \multicolumn{1}{c|}{\textbf{52.9}} &
  \multicolumn{1}{c|}{\textbf{23.4}} &
  \multicolumn{1}{c|}{42.3} &
  \textbf{54.0} \\ \hline
Trans-New Guinea &
  \multicolumn{1}{c|}{\textbf{24.9}} &
  \multicolumn{1}{c|}{\textbf{42.3}} &
  \multicolumn{1}{c|}{\textbf{56.2}} &
  \multicolumn{1}{c|}{\textbf{22.1}} &
  \multicolumn{1}{c|}{\textbf{42.1}} &
  \textbf{53.5} &
  \multicolumn{1}{c|}{22.2} &
  \multicolumn{1}{c|}{40.0} &
  \multicolumn{1}{c|}{53.2} &
  \multicolumn{1}{c|}{20.9} &
  \multicolumn{1}{c|}{41.2} &
  52.7 \\ \hline
\end{tabular}%
}
\caption{\label{tab:nllb_NT_completion}
BLEU, spBLEU, and chrF3 scores for \ac{NLLB}-600M fine-tuned models on \emph{\ac{NT} Completion} (without \emph{Related Language}) task and the \emph{\ac{NT} Completion} (with \emph{Related Language}) task.
}
\end{table}

\begin{acronym}[RNEMDS]
  \acro{NLP}{natural language processing}
  \acro{LWCs}{languages of wider communication}
  \acro{BT}{bible translation}
  \acro{NLLB}{No Language Left Behind}
  \acro{SMT}{statistical machine translation}
  \acro{NMT}{neural machine translation}
  \acro{NT}{New Testament}
  \acro{OT}{Old Testament}
  \acro{DT}{Deuterocanon}
  \acro{CV}{cross-validation}
  \acro{HMM}{hidden Markov model}
  \acro{USFM}{unified standard format markers}
\end{acronym}

\end{document}